\documentclass[letterpaper, 10 pt, conference]{ieeeconf} 
\IEEEoverridecommandlockouts

\usepackage{cite}
\usepackage{amsmath,amssymb,amsfonts}
\usepackage{algorithmic}
\usepackage{graphicx}
\graphicspath{{./figures}{./figures/new_experiments}}
\usepackage{hyperref}
\usepackage{textcomp}
\usepackage{xcolor}
\usepackage{subcaption}
\usepackage{booktabs}
\usepackage{tikz}
\usetikzlibrary{arrows.meta}
\usepackage{listings}
\usepackage{balance}

\def\BibTeX{{\rm B\kern-.05em{\sc i\kern-.025em b}\kern-.08em
    T\kern-.1667em\lower.7ex\hbox{E}\kern-.125emX}}
    
\begin{document}

\title{\huge VernaCopter: Disambiguated Natural-Language-Driven Robot via Formal Specifications}

\author{Teun van de Laar$^{1}$,~
Zengjie Zhang$^{1}$,~
Shuhao Qi$^{1}$,~
Sofie Haesaert$^{1}$,~and Zhiyong Sun$^{2}$
\thanks{This work was supported by the European project SymAware under grant No. 101070802, and by the European project COVER under grant No. 101086228.}
\thanks{$^{1}$T. Laar, Z. Zhang, S. Qi, and S. Haesaert are with the Department of Electrical Engineering, Eindhoven University of Technology,
PO Box 513, 5600 MB Eindhoven, Netherlands.
        {\tt\small \{t.a.v.d.laar@student.tue.nl, \{z.zhang3, s.qi, s.haesaert\}@tue.nl\}}}
\thanks{$^{2}$Z. Sun is with the College of Engineering, Peking University, Beijing, China. {\tt\small \{zhiyong.sun@pku.edu.cn\}}}
        }

\maketitle

\begin{abstract}
It has been an ambition of many to control autonomous robots using natural language (NL). The emergence of large language models (LLMs) has made this aspiration increasingly attainable. However, an LLM-powered system still suffers from the ambiguity inherent in NL and the uncertainty brought up by LLMs. This paper proposes a novel LLM-based robot motion planner, named \textit{VernaCopter}, with signal temporal logic (STL) specifications serving as a bridge between NL commands and specific task objectives. The rigorous and unambiguous nature of formal specifications allows the planner to generate high-quality and highly consistent trajectories to guide the motion control of a robot. Compared to a conventional NL-prompting-based planner, the proposed VernaCopter planner is more stable and reliable due to lower levels of ambiguity. Two challenging experimental scenarios have validated its efficacy and advantage, implying its potential for NL-driven robots.
\end{abstract}


\section{Introduction}\label{sec:intro}

The control of robotic systems typically requires clearly defined commands or specifications prescribed by experts. Controlling robots using natural language (NL) is still an open and challenging topic~\cite{ahn2018interactive}. This is mainly due to the ambiguity of NL, which is attributed to its implicit and complex semantics~\cite{zemni2024comparative}. Conventionally, language-driven robot control has been achieved by introducing an NL processing (NLP) unit that translates human language into concrete tasks~\cite{kahuttanaseth2018commanding, johri2021natural}. The recent rise of Large language models (LLMs) like ChatGPT \cite{gpt3} and Gemini \cite{geminiteam2023gemini} provides an efficient manner to control robots from NL commands directly. This offers a practical way for non-expert users to control robots using NL~\cite{naveed2024comprehensive}. Facilitated by a transformer architecture, LLMs can capture the relationships between words and phrases in a sentence by leveraging the attention mechanism~\cite{xu2024survey}. Therefore, LLMs can understand the context of an NL command and reason about its logic, bridging  the gap between ambiguous language and explicit specifications~\cite{zhang2024llm}. Successful instances of language-controlled robots include manipulation~\cite{vemprala2023chatgpt}, navigation~\cite{Chen2023}, and conversation services~\cite{wang2024llm}. A critical technical point of ensuring success in LLM-enabled robot control is designing an appropriate prompting scheme, for which prompt engineering provides useful guidelines~\cite{arawjo2024chainforge}. A survey on LLM-controlled robots can be referred to in~\cite{zhao2024survey}.

\begin{figure}[htbp]
\centering
    \includegraphics[width=\linewidth]{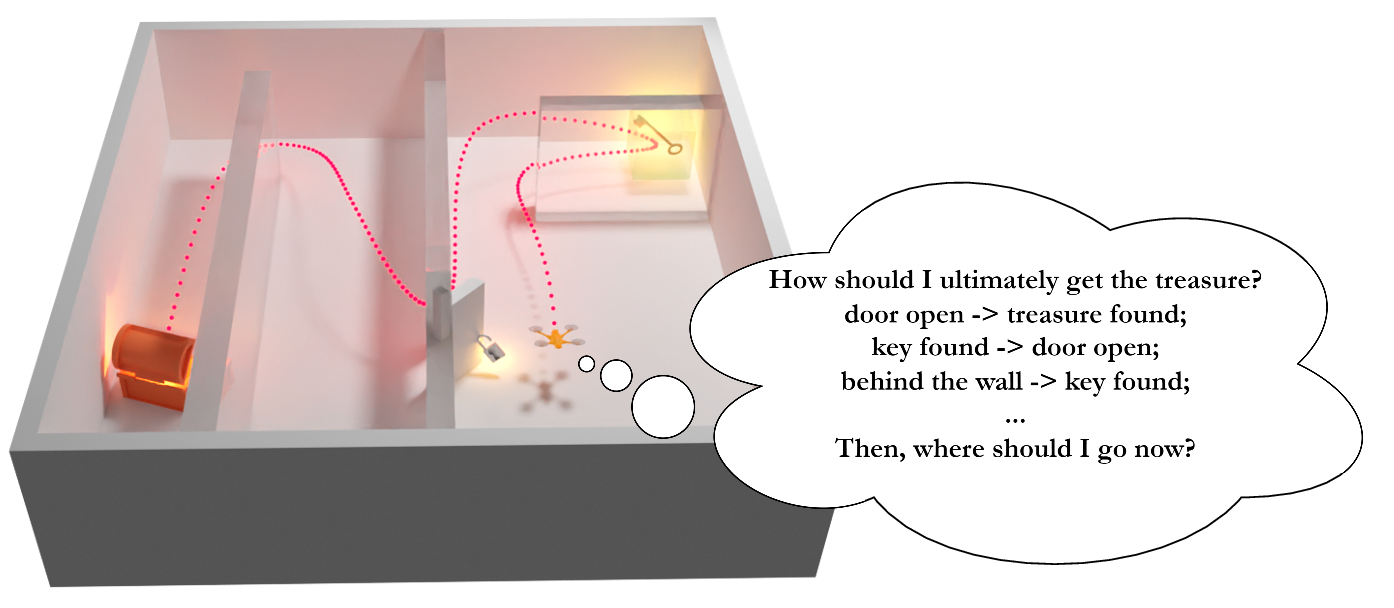}
    \caption{A treasure-hunt quadcopter with interdependent subtasks.}
\label{fig:cop_scenarios}
\end{figure}

On the other hand, LLMs are also known for their drawbacks in ambiguity and uncertainty~\cite{liu2024uncertainty, gao2023ambiguity}. Specifically, LLMs are sensitive to the input prompts. LLMs can return distinguished outputs for the same prompts at different instances, potentially leading to additional uncertainty when used for robot control. An additional source of ambiguity and uncertainty is the spatial and numerical reasoning limitations that prevent the LLM from correctly reconstructing the logical interdependence among different subtasks in a complex task~ \cite{wake2023chatgpt, valmeekam2022large}. It has been argued that LLMs are less stable and reliable for complicated robot motion planning tasks than simple ones~\cite{liu2024enhancing}. Take the scenario illustrated in Fig.~\ref{fig:cop_scenarios} as an example, where a quadcopter needs to retrieve a treasure from a chest in a room behind a locked door, which can be opened with a key. This renders a complex task containing a sequence of three interdependent subtasks, namely \textit{key fetching}, \textit{door unlocking}, and \textit{box opening}. In the rear part of this paper, we will use a case study to show that an LLM does not always succeed in this task since it cannot properly reconstruct the logical relations between these subtasks from the NL prompts.

\begin{figure}[htbp]
\centering
    \centering
    \includegraphics[width=\linewidth]{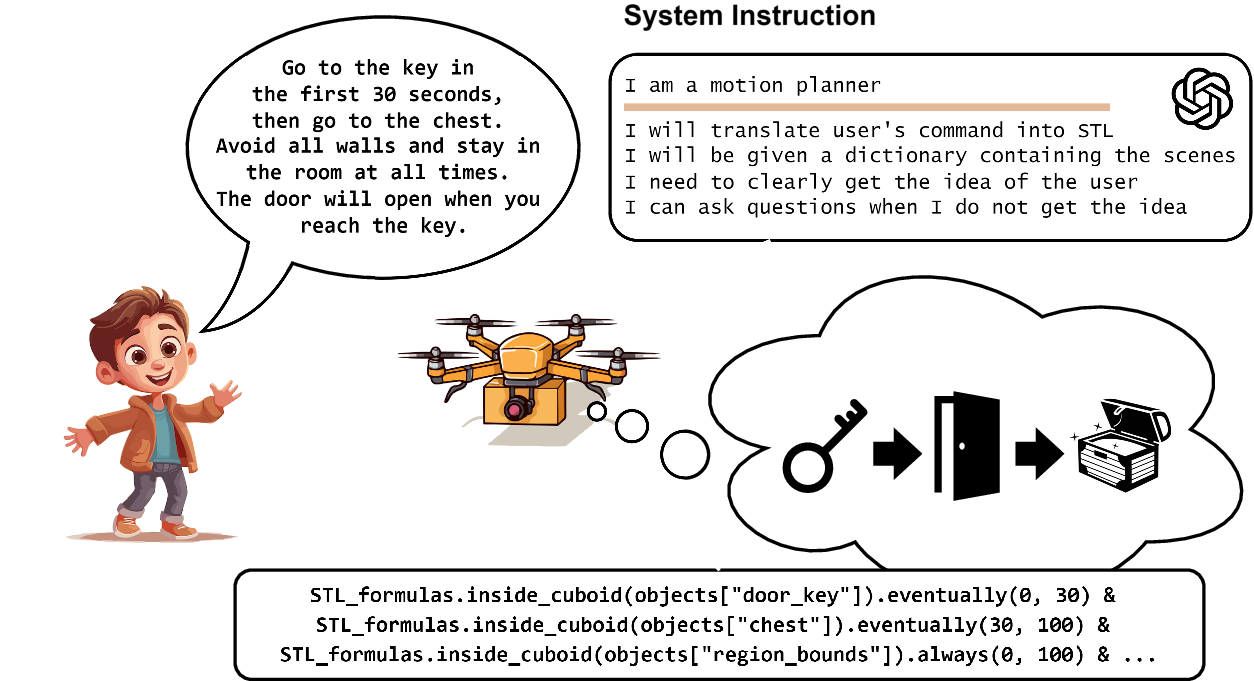}
    \caption{Facilitating LLM-based robots with STL specifications.}
\label{fig:inquiry}
\end{figure}

Many recent works have been dedicated to promoting the reliability of LLMs for robot control tasks. Some of them specifically focus on refining prompting techniques for action sequences using Pythonic programming structures, leveraging the vast amount of Python data used to train ChatGPT~\cite{vemprala2023chatgpt, singh2023progprompt, wake2023chatgpt, ding2023task}. Other research addresses the limitations of code/action sequence-based methods. RobotGPT \cite{10412086} trains a separate agent on task plans generated by ChatGPT to increase reliability. To improve the scalability of LLM-based frameworks, SayPlan \cite{rana2023sayplan} employs a three-dimensional scene graph to search the relevant parts of the scene semantically. The executability of proposed action sequences is addressed in \cite{huang2022language, ahn2022do, lin2023text2motion}. Other frameworks include visual information to ground the LLM visually in real-world environments \cite{wu2023tidybot, huang2022inner, Bucker2022}. However, guaranteeing optimality and safety for code/action sequence-based frameworks is impractical in general. Meanwhile, formal specifications like linear-time temporal logic (LTL) and signal temporal logic (STL) play an important role in ensuring safety and risk restriction for robot control and motion planning~\cite{kloetzer2007temporal, lindemann2019coupled, qi23cdc}. They provide a generic framework with abstract semantics to describe robotic tasks, providing a potential approach to improving the reliability of LLM-centered systems. As illustrated in Fig.~\ref{fig:inquiry}, a possible solution is to let the LLM generate STL specifications as a bridging language to guide the motion generation, instead of explicit robot trajectories as performed by a conventional LLM-centered planner. The exactness inherent in formal specifications ensures a structured approach to addressing the intrinsic ambiguity of NL.

We design a novel LLM-centered motion planning system for a drone-like robot, also known as \textit{VernaCopter}, dedicated to complex navigation tasks. The VernaCopter planner aims to allow non-expert users to control drone-like robots using NL commands without specific guidance from experts. The prefix “\textit{Verna}”, derived from the word “\textit{vernacular},” represents the natural conversations between users and the assistant, and the suffix “\textit{Copter}” signifies the application to quadcopters. The planner leverages STL as a bridge between the NL commands and the concrete task objectives, in a way to catalyze the efficiency and reliability of the system by reducing the ambiguity of NL and the uncertainty of the LLM agent. The planner has a compact structure, allowing for easy implementation. In the meantime, it provides several function modes that can be customized for different levels of efficiency and complexity. Experimental studies have verified that the proposed planner can significantly increase the success rate for complex motion planning tasks compared to the conventional prompt-based LLM-centered motion planner. Our main contributions are threefold:

\begin{enumerate}
    \item A novel easy-to-implement STL-enabled LLM-centered robot motion planner with reduced ambiguity.
    \item Effective prompting techniques ensuring a high success rate.
    \item The design of assistant LLM agents to promote the planning performance.
\end{enumerate}

The remaining contents of this paper are organized as follows. In Sec.~\ref{sec:Preliminaries}, we elaborate on the related work on two important technologies to support this work, namely prompt design and formal specifications. Sec. \ref{sec:Architecture} interprets the design of the VernaCopter planner system and Sec. \ref{sec:Prompt_engineering} introduces how we prompt the system. Sec. \ref{sec:Experimental_studies} uses two case studies to validate the system's efficacy. Sec.~\ref{sec:Discussion} elaborates on the strengths, limitations, and possible extensions of our system. Finally, Sec. \ref{sec:Conclusions} concludes the paper.



\textit{\textbf{Video Demonstration:}} A video demonstrating the VernaCopter planner is at {\small \url{https://youtu.be/bAVh3qgUIUA}}.

\section{Related Work}\label{sec:Preliminaries}


\subsection{Specifying Tasks via Prompts}

In NLP, where the model's weights are not affected by each request, the quality and specificity of generated outputs are highly dependent on the formulation of the prompting inputs. These prompts encompass both general instructions provided upfront, describing the LLM's tasks and limitations, and specific prompts users provide during interaction. In designing the system, the general task descriptions for the LLM are prompt-based, so using appropriate prompts is critical for the system's behavior and performance. However, applying appropriate prompting techniques is also essential for the system's users. Several techniques exist for effective prompting in NLP. Some helpful information on prompting structures using ChatGPT techniques is presented in \cite{white2023prompt, ekin2023prompt}. 
The work in~\cite{shi2024yell} argues that long-horizon tasks possibly lead to more uncertainties which can be improved by human feedback, rendering a conservation-based prompting scheme. In~\cite{liu2024leveraging}, a prompt-based scheme is developed to induce an LLM to point out the successive subtasks. There are also attempts dedicated to promoting the precision of LLM using additional machine learning components, such as text encoders~\cite{yuan2024rag}.
In~\cite{gu2023systematic}, visual-language models are studied to provide a solid foundation for potential future research concerning applying LLMs to robotics. Moreover, it categorizes prompting techniques into the following types which are partially generated using ChatGPT and are carefully reviewed and refactored.

\begin{itemize}
\item \textbf{Task instruction prompting} describes a task explicitly in as detailed as necessary to define the task thoroughly.
\item \textbf{In-context learning} relies on several examples closely related to the task to generalize to new tasks.
\item \textbf{Chain-of-thought prompting} encourages a model to provide intermediate reasoning steps rather than returning a single result as a response to an instruction. This method can enhance the quality of the response significantly \cite{wei2023chainofthought}.
\end{itemize}

These prompting techniques have different strengths. \textit{Task instruction prompting} can be used to describe tasks and limitations. \textit{In-context learning} is particularly useful when using a pre-trained LLM, but the model is not specifically trained in the context in which it operates. In this case, examples can provide details on using STL appropriately. Finally, \textit{chain-of-thought prompting} can increase the accuracy of outputs, especially when reasoning about complex problems.

Most LLM-based robot control frameworks as mentioned above directly generate robot trajectories from NL-based prompts from the user, leading to possible uncertainty due to the ambiguous nature of NL. This is due to the disadvantage of LLM when reasoning about numerical and geometrical relations between the robot states and the task space~\cite{sun2024determlr}. As a result, pure NL-prompted LLM may not correctly interpret the correct logical relation between the subtasks of a complex task, potentially leading to failures. Next, we elaborate on the related work on formal specifications which provide an abstract and rigorous interface for task description.

\subsection{Specifying Tasks via Formal Specifications}

Different from NL, formal specifications provide precise descriptions for robotic tasks~\cite{Plaku2015MotionPW}. Formal language-based motion planning is a well-studied subject, and motion plans generated from formal language are optimal and safe by design~\cite{da2016formal}. In this sense, a straightforward approach is to use an LLM model to translate tasks specified in NL into a formal specification which is then synthesized for robot motion planning~\cite{Chen2023, Xie2023, Liu2023}. 

Nevertheless, introducing formal language in LLM-based robotics frameworks brings additional challenges. Correctly generating formal language specifications from NL is non-trivial due to the ambiguous nature of NL \cite{8972130}. Furthermore, LLMs are black box systems, meaning their outputs are not traceable and could be inconsistent when given the same input. The syntax of generated specifications may conflict with the original, and the specified task and the corresponding generated formal language specification could be semantically misaligned. Additionally, while the code/action sequence-based methods could leverage LLMs' extensive code training, LLMs are not directly trained on formal language data quantities of the same magnitude. 

The challenges of syntactic errors and semantic misalignments of formal specifications can be addressed by applying LLMs to identify and correct faulty results. Iterative checking and correcting modules are used in \cite{Chen2023}, correcting syntax using a rule-based approach and aligning semantics by re-prompting the LLM with the original task description. The challenge of missing extensive training in using formal specifications can be addressed by deploying prompting techniques. Appropriate prompting can provide information to pre-trained LLMs to reason about and construct STL specifications.

\section{VernaCopter Planner System}\label{sec:Architecture}

This section introduces the composition of the VernaCopter planner system. The overall architecture and workflow of the system are first interpreted, followed by a detailed elaboration of its specific components.

\subsection{Overall Architecture}\label{sec:oa}

\begin{figure}[htbp]
    \noindent
    \hspace*{\fill}
    \begin{tikzpicture}[scale=1, font=\small]

        \definecolor{sorange}{RGB}{255, 205, 153}
        \definecolor{sblue}{RGB}{204, 221, 255}
        \definecolor{sred}{RGB}{255, 235, 229}
        \definecolor{sgreen}{RGB}{229, 255, 229}
        \definecolor{syellow}{RGB}{255, 255, 245}
        \definecolor{spurple}{RGB}{255, 229, 255}
        \definecolor{sgray}{RGB}{242, 242, 242}

        \node[minimum width=2.2cm, minimum height=6.4cm, draw, dashed] (ui) at (0cm, 1.05cm) {};

        \node[minimum width=5.0cm, minimum height=6.4cm, draw, dashed] (fw) at (4.2cm, 1.05cm) {};

        \node[minimum width=1.8cm, minimum height=1.8cm, rounded corners=0.2cm, text width=1.6cm, draw, thick, fill=sred, align=center] (user) at (0cm, 2.3cm) {\textbf{User \\ \vspace{0.4cm}  \includegraphics[width=0.6cm]{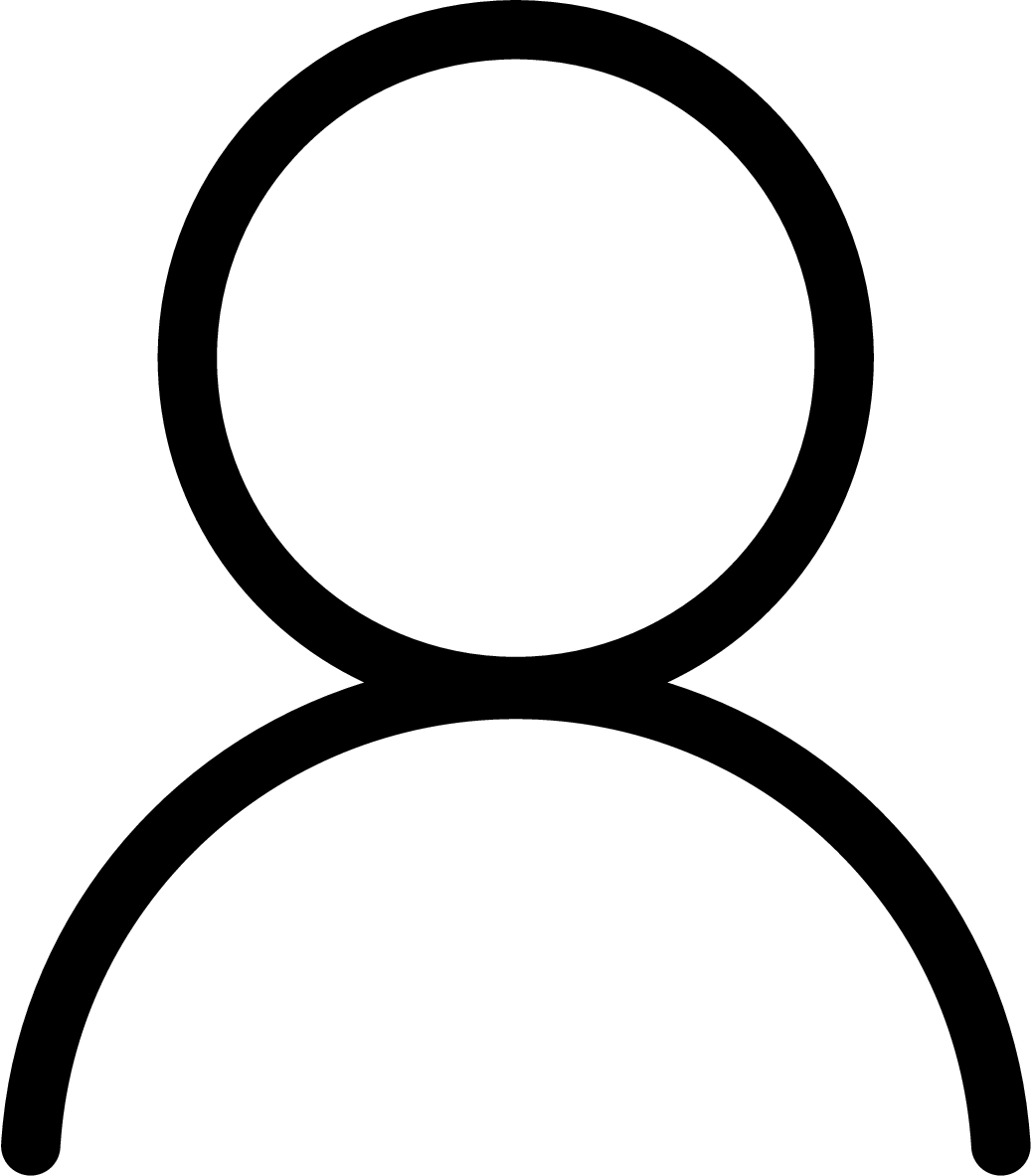}}};

        \node[minimum width=1.8cm, minimum height=1.8cm, rounded corners=0.2cm, text width=1.6cm, draw, align=center, fill=sgreen] (pa) at (2.8cm, 2.3cm) {\footnotesize \textbf{Planning \\ Assistant \\ (PA) \\ \includegraphics[width=0.6cm]{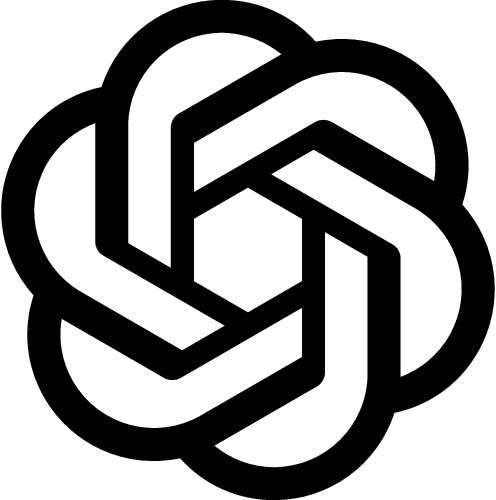}}};

        \node[minimum width=1.8cm, minimum height=1.8cm, rounded corners=0.2cm, text width=1.6cm, draw, fill=syellow, align=center] (sem) at (5.6cm, 2.3cm) {\footnotesize \textbf{Semantics \\ Checker (SemCheQ) \includegraphics[width=0.6cm]{ChatGPT-Logo.eps}}};

        \node[minimum width=1.8cm, minimum height=1.8cm, rounded corners=0.2cm, text width=1.6cm, draw, fill=syellow, align=center] (syn) at (2.8cm, 0cm) {\footnotesize \textbf{Syntax \\ Checker (SynCheQ) \includegraphics[width=0.6cm]{ChatGPT-Logo.eps}}};

        \node[minimum width=1.8cm, minimum height=1.8cm, rounded corners=0.2cm, text width=1.6cm, draw, align=center, thick, fill=sgray] (vis) at (0cm, 0cm) {\textbf{Visualizer}\vspace{0.4cm}  \includegraphics[width=0.6cm]{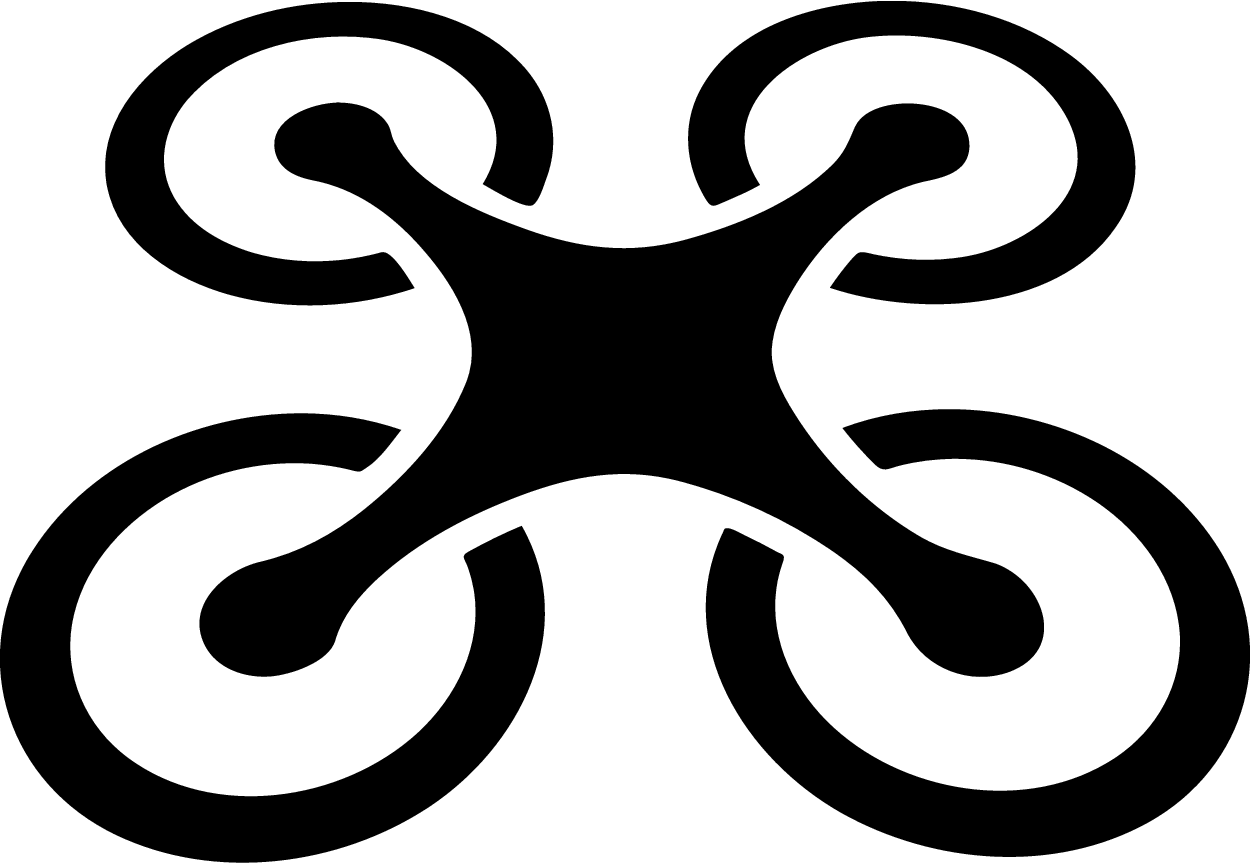}};

        \node[minimum width=1.8cm, minimum height=1.8cm, rounded corners=0.2cm, text width=1.6cm, draw, align=center, thick,fill=sblue] (traj) at (5.6cm, 0cm) {\footnotesize \textbf{Path \\AnalyZer\\(PAZ)}  \includegraphics[width=0.9cm]{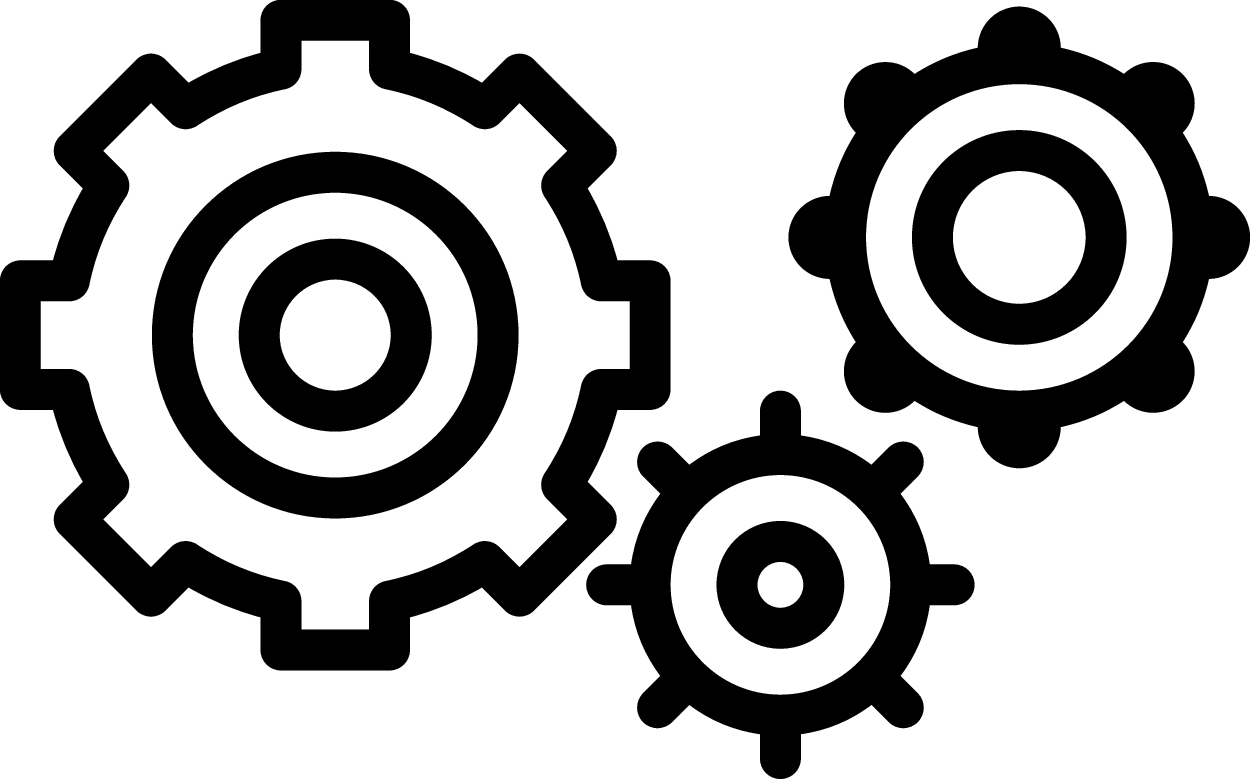}};

        \node[minimum width=0.6cm, minimum height=0.6cm, draw, thick, dotted, align=center] (opt) at (2.8cm, -1.7cm) {\textbf{Optimizer}};

        \node[align=left, anchor=north west] () at (ui.north west) {\textbf{User Interface}};

        \node[align=left, anchor=north west] () at (fw.north west) {\textbf{VernaCopter Planner}};

        \draw [->, >=Stealth, thick] (user) -- node[anchor=south]{\footnotesize NL} (pa);
        \draw [->, >=Stealth, thick] (pa) -- node[anchor=west]{\footnotesize STL formula} (syn);
        \draw [->, >=Stealth, thick] (sem) -- node[anchor=south]{\footnotesize Advice} (pa);

        \draw [->, >=Stealth, thick] (user.north) -- ([yshift=0.5cm] user.north) -- node[pos=0.5, anchor=north, align=center, fill=white]{\footnotesize Task Description} ([yshift=0.5cm] pa.north) -- (pa.north);
        \draw [->, >=Stealth, thick] ([yshift=0.5cm] pa.north) -- ([yshift=0.5cm] sem.north) -- (sem.north);

        \draw [->, >=Stealth, thick] ([xshift=-0.4cm] syn.south) -- node[anchor=east]{\footnotesize Checked STL} ([xshift=-0.4cm] opt.north);

        \draw [<-, >=Stealth, thick] ([xshift=0.4cm] syn.south) -- node[anchor=west]{\footnotesize Error} ([xshift=0.4cm] opt.north);

        \draw [->, >=Stealth, thick] ([yshift=-0.2cm] opt.east) -- node[pos=0.5, anchor=south]{\footnotesize Path (waypoints)} ([yshift=-1cm] traj.south) -- (traj.south);

        \draw [->, >=Stealth, thick] ([yshift=-0.2cm] opt.west) -- node[pos=0.5, anchor=south]{\footnotesize Path (waypoints)}([yshift=-1cm] vis.south) -- (vis.south);

        \draw [->, >=Stealth, thick] (traj) -- node[anchor=west]{\footnotesize Texts} (sem);

        \draw [->, >=Stealth, thick, dotted] (vis) -- node[anchor=west]{\footnotesize Vision} (user);

        \node[] () at ([xshift=-0.3cm, yshift=-0.7cm] fw.west) {\reflectbox{\includegraphics[width=0.5cm, angle=45]{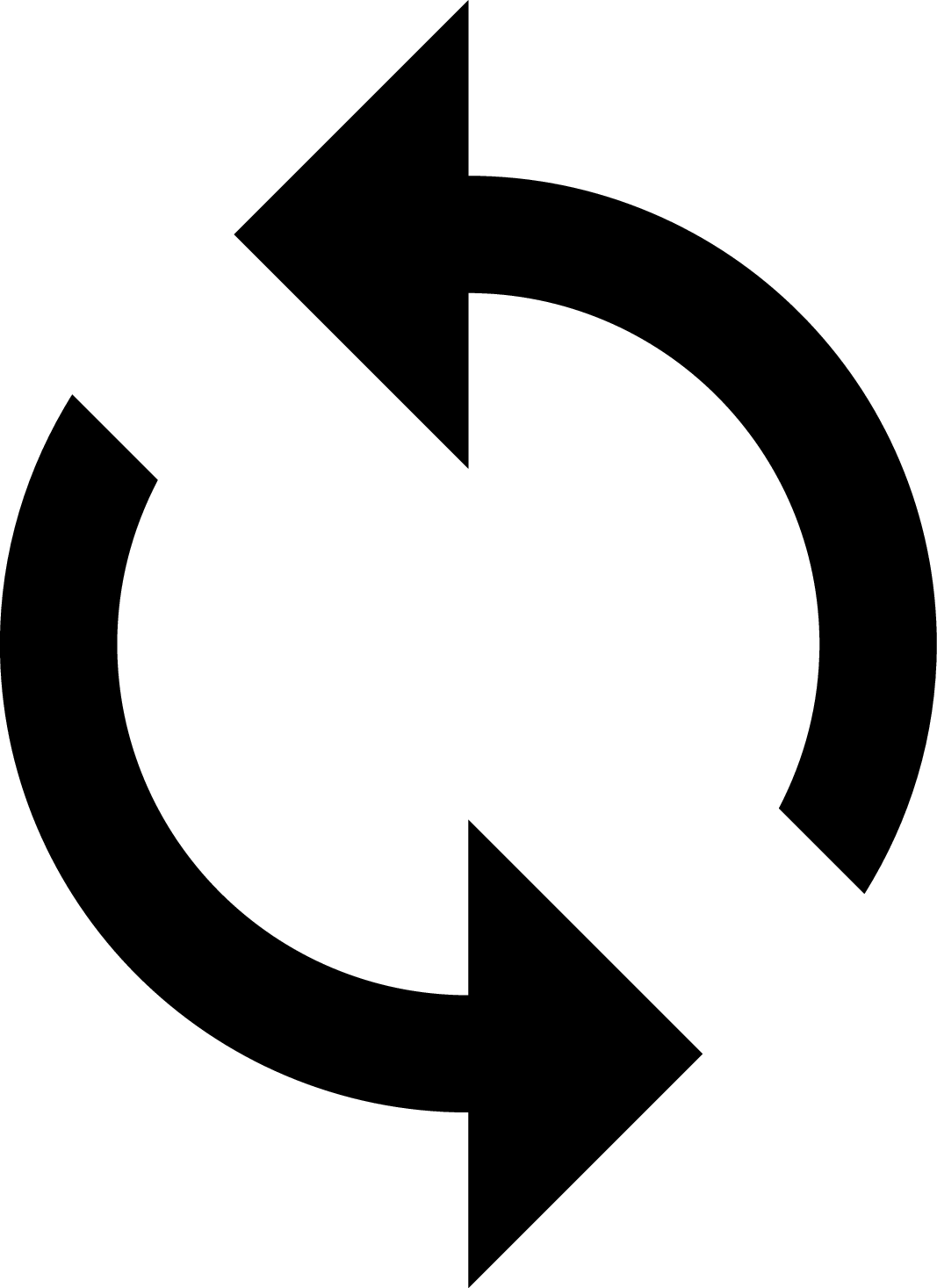}}};
        \node[] () at ([xshift=-0.3cm, yshift=-0.7cm] fw.west) {\color{red} \small \textbf{1}};
        
        \node[] () at ([xshift=-0cm, yshift=-0.7cm] fw.center) {\includegraphics[width=0.5cm, angle=45]{loop-5.eps}};
        \node[] () at ([xshift=-0cm, yshift=-0.7cm] fw.center) {\color{red} \small \textbf{3}};
        
        \node[] () at ([yshift=-0.25cm] syn.south) {\includegraphics[width=0.5cm, angle=45]{loop-5.eps}};
        \node[] () at ([yshift=-0.25cm] syn.south) {\color{red} \small \textbf{2}};


    \end{tikzpicture}
    \hspace{\fill}
    \caption{The VernaCopter planner architecture.}
    \label{fig:architecture}
\end{figure}

The overall architecture of the VernaCopter planner is illustrated in Fig. \ref{fig:architecture}. An LLM as a planning assistant (PA) serves as a human-robot interface, translating the user command in NL into STL specifications. The detailed definition of STL can be referred to in Appx.-A. The user command includes a task description as one-shot system instruction and iterative prompts in the form of question-and-answer (Q\&A) conversations. Another two LLM-based checkers are used to validate the correctness of the specifications, from the syntactical and semantic perspectives, respectively. Then, the system model with the generated specifications is synthesized using an optimizer, with the resulting trajectories analyzed for automatic improvement and visualized for manual inspection.

The workflow of the VernaCopter planner is summarized as three loops, marked as arrowed circles with numbers in Fig.~\ref{fig:architecture}. Loop 1 involves 
the PA generating an STL formula according to the user prompts. A Syntax Checker (SynCheQ) checks the correctness of the STL formula and gives the corrected formula to an optimizer. The solved path solution is visualized in a simulation environment, such as Pybullet, for the user to inspect. In the case of a solving error, such as infeasibility, the optimizer returns an error flag to SynCheQ to fine-tune the STL formula, forming Loop 2. In Loop 3, a path analyzer (PAZ) is designed to describe a generated path in text. According to its output, a Semantics Checker (SemCheQ) provides improvement advice to the PA taking into account the task description.

In general, 
Loop 1 is dedicated to generating an STL specification from NL, Loop 2 guarantees the syntactic correctness and feasibility of the generated specification, and Loop 3 ensures semantic alignment of the path with the task description. Different from the existing loop-based prompting schemes~\cite{Chen2023}, the VernaCopter planner only uses loops to improve its performance, instead of aiming at a converging generation. In Sec.~\ref{sec:Experimental_studies}, we will showcase that properly designed one-shot prompts can achieve decent planning performance without performing loop-based iterations.

\subsection{Planning Assistant (PA)}

The PA is an LLM agent that translates a user-defined task formulated in NL to an STL specification. In this paper, we adopt a pre-trained GPT model from OpenAI \cite{gpt3, openai2024gpt4}, namely GPT-4o. The LLM is instructed on its general task and is provided with a library of available STL functions and operators and several examples of correct uses of STL. In this basic form of the architecture, the syntactic correctness of the specification is ensured by providing examples of user inputs and desirable outputs. The examples provided partially ensure the semantic alignment of the task and the generated STL specification. An abstraction of the scene is provided to the LLM, consisting of names of objects and their corresponding bounding boxes. In Sec. \ref{sec:Prompt_engineering}, we will interpret how to catalyze the performance of the LLM-based PA by leveraging prompt-engineering techniques. 


\subsection{Syntax Checker and Optimizer}\label{sec:Extended_architecture}

The PA may not always generate a meaningful STL specification. Instead, LLM may produce syntactic errors due to the lack of correctness guarantees, leading to potential failure of the robot task. In this paper, we employ an LLM as a Syntax Checker (SynCheQ) to correct the possible syntactic errors of the generated STL formula. The SynCheQ is prompted with initial instructions that contain similar information to the PA. However, the general explanation and examples are targeted more toward syntax checking. The LLM is explicitly instructed not to change the semantics of the specification. SynCheQ either revises the STL if a syntactic error is found or returns the originally correct STL. The optimizer synthesizes the returned STL via a model-checking process using the Python package stlpy \cite{kurtz2022mixed}. The details of the model-checking and synthesis processes can be referred to in Appx.-B.

\subsection{Visualizer}\label{sec:Interface_with_LLM_and_visualization}

Once the final trajectories of an autonomous robot are generated, the behavior of VernaCopter is visualized using \textit{gym-pybullet-drones} \cite{panerati2021learning}, a drone simulation environment based on OpenAI's Gym \cite{1606.01540}. The motion of the drone is controlled using a pre-tuned PID controller implemented in \textit{gym-pybullet-drones}. 
A fixed control frequency is used to control the drone in the simulation. To make the drone follow the path defined by the waypoints, the target waypoint is switched to the next one in the sequence with this same frequency. The user can accept or reject the proposed path based on this information. Whenever a path is rejected, the user can correct the reasoning and output from the LLM. An expert on STL or a technical user can use the specification to correct any semantic misalignment between the specification and the task. However, the user is not required to know STL. The LLM can assist the user by explaining the details of the generated specification when requested. Based on the new instructions, a new path is generated. This process forms Loop 1 which is repeated until the task is satisfied.

\subsection{Semantics Checker and Path AnalyZer (PAZ)}\label{sec:Semantics_module}

Although the correctness and feasibility of the generated STL are ensured by the SynCheQ in Loops 1 and 2, it may not align with the task description. In many cases, it may not precisely characterize the interdependence among different subtasks. Thus, a semantic alignment checker (SemCheQ) is deployed to correct these oversights. SemCheQ is an LLM agent that judges whether the task description and the generated path are semantically identical. To facilitate this, a PAZ is designed to analyze the generated path and automatically generate a textual description of this path, which is sent to the SemCheQ. Specifically, the PAZ checks which regions or objects are traversed throughout the generated path. The exported textual description contains as many elements as there are objects in the scene. For each object, it describes whether the drone \textit{a)} stays in that object's bounding box at all times, \textit{b)} is outside the bounding box at all times, or \textit{c)} is inside the bounding box of the object during some specific times only. The specific time steps are also returned to SemCheQ in the latter case. When the task description and the generated path are not semantically identical, SemCheQ forwards an advice message to the PA to improve the specification.


\section{Leveraging Instructive Prompting}\label{sec:Prompt_engineering}

The performance of an LLM-centered system highly depends on the quality of prompts. Similar to other LLM-centered systems mentioned in Sec.~\ref{sec:Preliminaries}, the performance of the VernaCopter planner can be catalyzed by properly designing these prompts. The proposed VernaCopter planner supports one-shot prompting, already achieving decent performance for small robot motion planning tasks. The user directly provides the command in NL to the PA. When starting the program, the LLM is automatically prompted with general instructions. These instructions contain information about the general task, the environment, available functions, operators, the syntax to construct the STL specification, how to interact with the user, and examples of appropriate responses. The details of these instructions are shown in Appx.-C.

The planner can be used in a conversation mode, allowing the user to continually instruct the PA and confirm whether it receives all necessary information for the given task. The LLM decides when it has gained enough information to generate a specification, and a set of waypoints is generated using the specification. The chain-of-thought reasoning generated by the LLM is printed to the user, and the generated waypoints are visualized in an environment abstraction for validation. The user is not obligated to provide the full task or all restrictions directly, as the assistant is instructed to elucidate missing details collaboratively. The user can converse as they see fit in response to the LLM's outputs. 

The proposed prompting instructions are not specific to a particular scenario but can be customized for various scenarios. To this end, the objects and regions are provided independently for each environment. Task instruction prompting is used to explain the general nature of the task and in-context learning is applied in a few-shot fashion, i.e., by providing several examples of prompts and appropriate responses. The LLM is encouraged to use chain-of-thought reasoning to improve the correctness of the outputs. 

\section{Experimental studies}\label{sec:Experimental_studies}

\begin{figure}[htbp]
\centering
\begin{subfigure}{0.46\linewidth}
    \includegraphics[width=\linewidth]{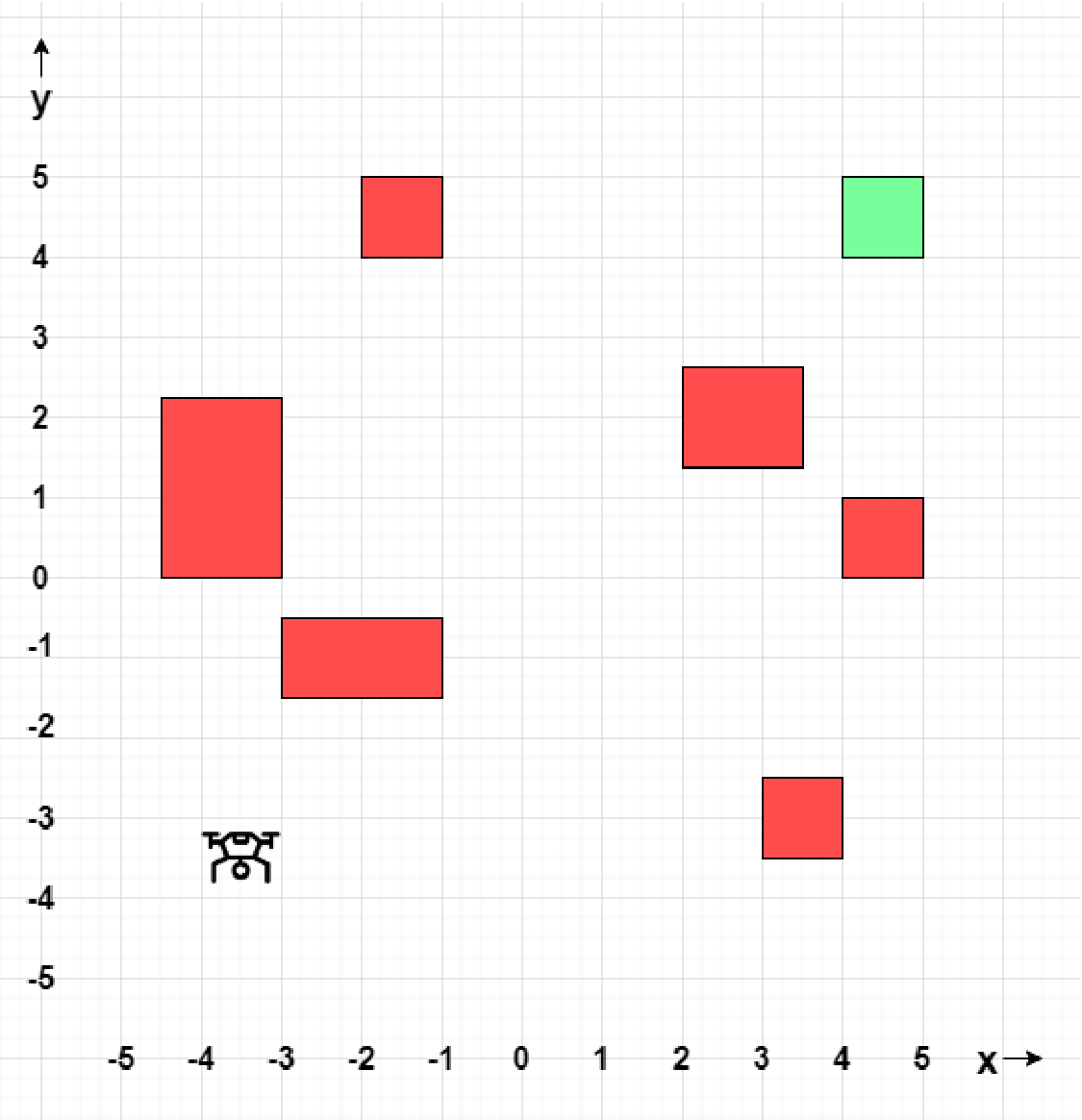}
    \caption{Reach-and-avoid (R\&A).}
    \label{fig:Reach_avoid}
\end{subfigure}
\hfill
\begin{subfigure}{0.46\linewidth}
    \includegraphics[width=\linewidth]{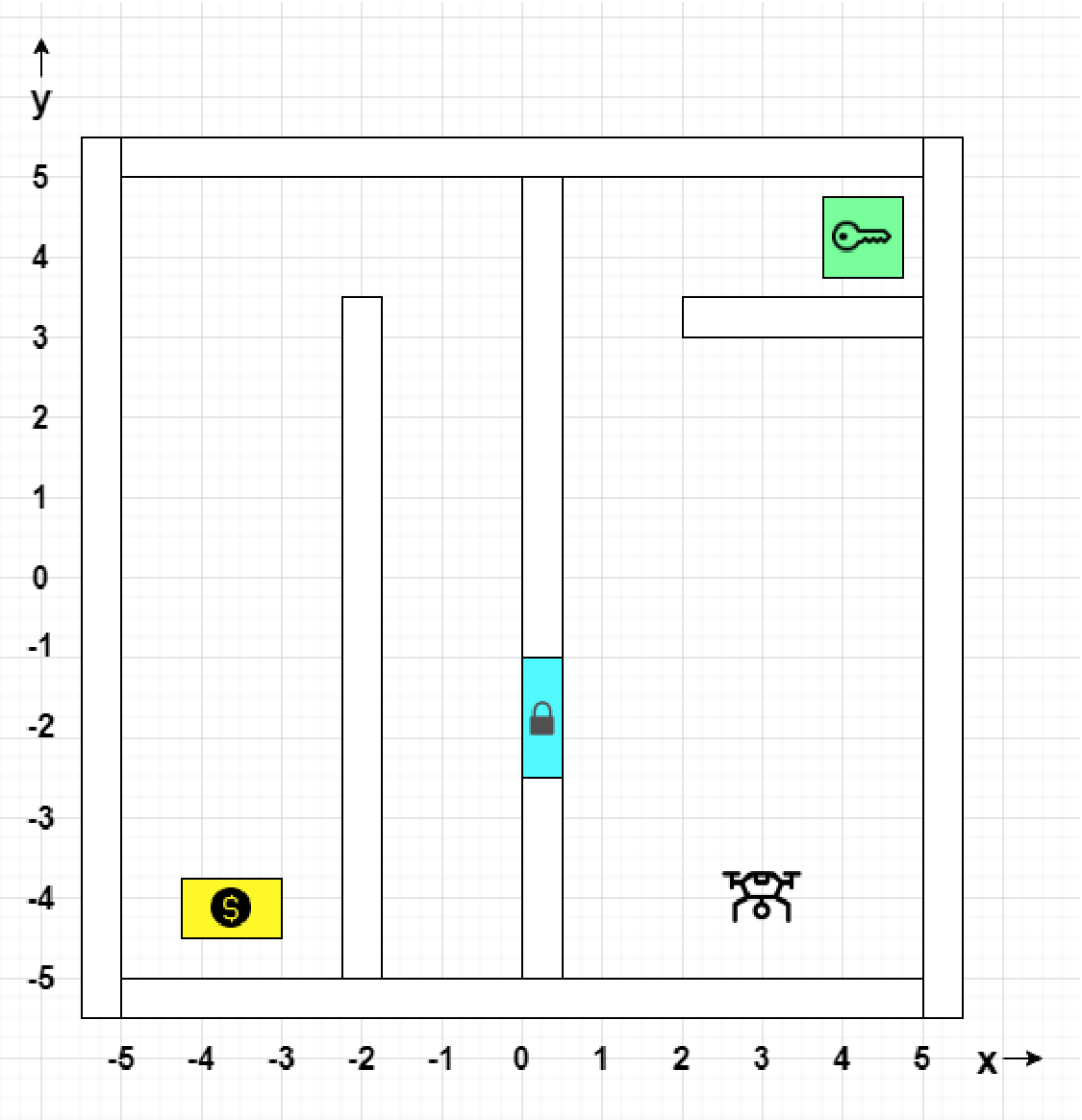}
    \caption{Treasure hunt.}
    \label{fig:treasure_hunt}
\end{subfigure}
\caption{The 2D views of the two experimental scenarios.}
\label{fig:scenarios}
\end{figure}

We use experimental studies to showcase how the VernaCopter planner ensures a high success rate for a robot task with reduced ambiguity, showing an advantage over a conventional prompting scheme without leveraging formal specifications. Two representative scenarios with different complexity, namely a reach-and-avoid (R\&A) task~\cite{paranjape2015motion} and a treasure hunt task are considered. Both scenarios are designed in 3D space. Their top views are illustrated in Fig.~\ref{fig:scenarios}. Each case contains a comparison study with a conventional NL-based planner. 

\subsection{Case I: Reach-and-Avoid (R\&A)}\label{sec:case1}

The R\&A scenario contains one goal (green) and several obstacles (red) of different sizes, as shown in Fig.~\ref{fig:Reach_avoid}. The VernaCopter is required to ultimately reach the goal from the opposite corner without colliding with any obstacles. It renders an essential but representative motion planning case that does not contain any subtasks. We intentionally involve a dense set of obstacles to increase the task's difficulty. This scenario is designed to assess the basic functionality of the Vernacopter for a simple task in a crowded environment. The result reflects the planner's stable and reliable performance abilities for a small but difficult problem.

For both the VernaCopter and the conventional NL-based planners, we use the one-shot prompting template in Appx.-C to deliver the task description to the PA. The input is a brief command in NL: ``\textit{Reach the goal while avoiding all obstacles.}". The conventional NL-based planner follows a similar structure to the VernaCopter planner, but the conventional PA generates robot paths directly, rather than STL specifications. For a fair comparison, the conventional planner is also equipped with a similar SemCheQ to ensure alignment with the task description. We perform multiple trials (50 for VernaCopter and 45 for conventional) to inspect whether they generate successful paths.

\begin{figure}[htbp]
\centering
\begin{subfigure}{0.48\linewidth}
    \includegraphics[width=\linewidth, trim={2cm 5cm 2cm 6cm},clip]{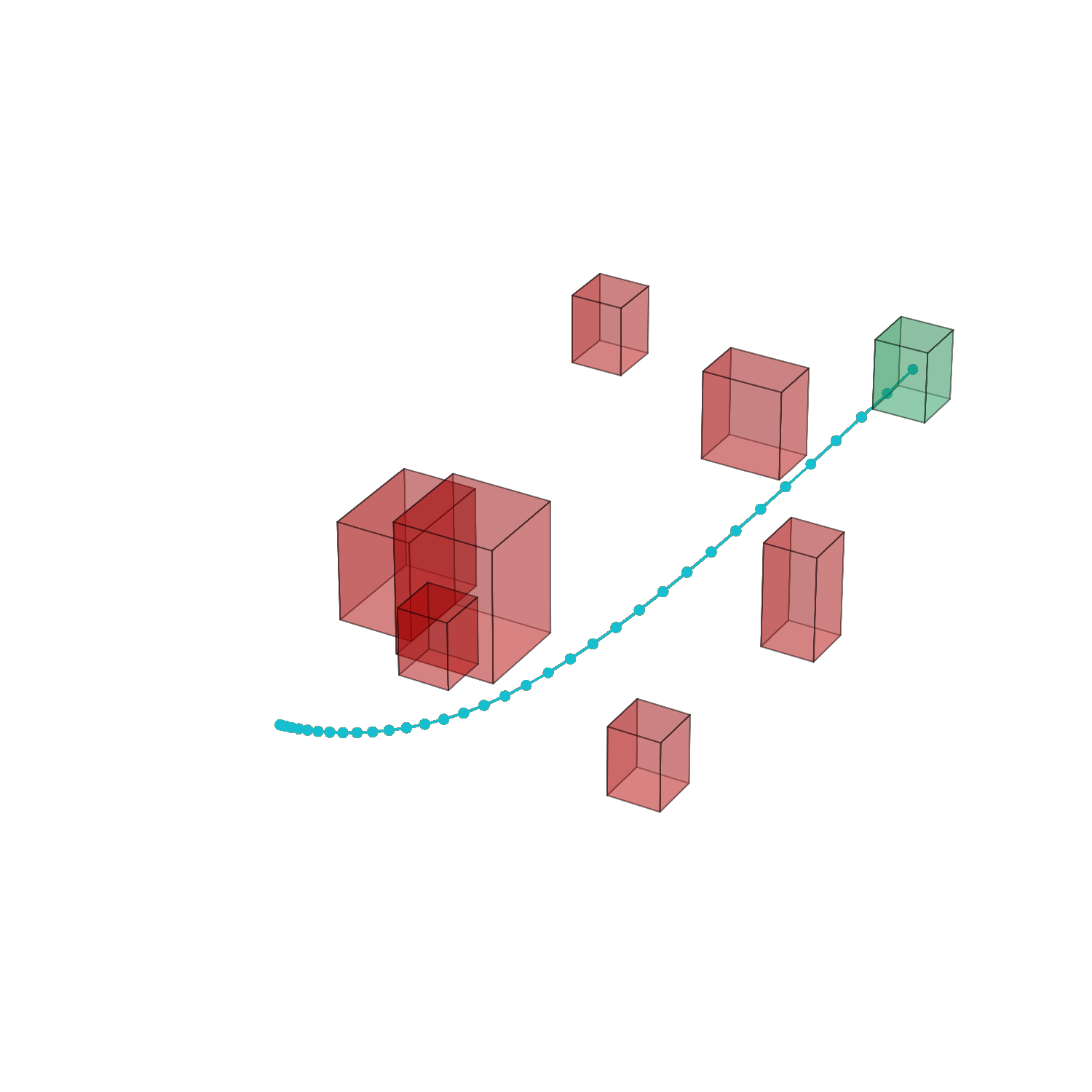}
    \caption{VernaCopter planner}
    \label{fig:randa_vcp}
\end{subfigure}
\begin{subfigure}{0.48\linewidth}
    \includegraphics[width=\linewidth, trim={2cm 5cm 2cm 6cm},clip]{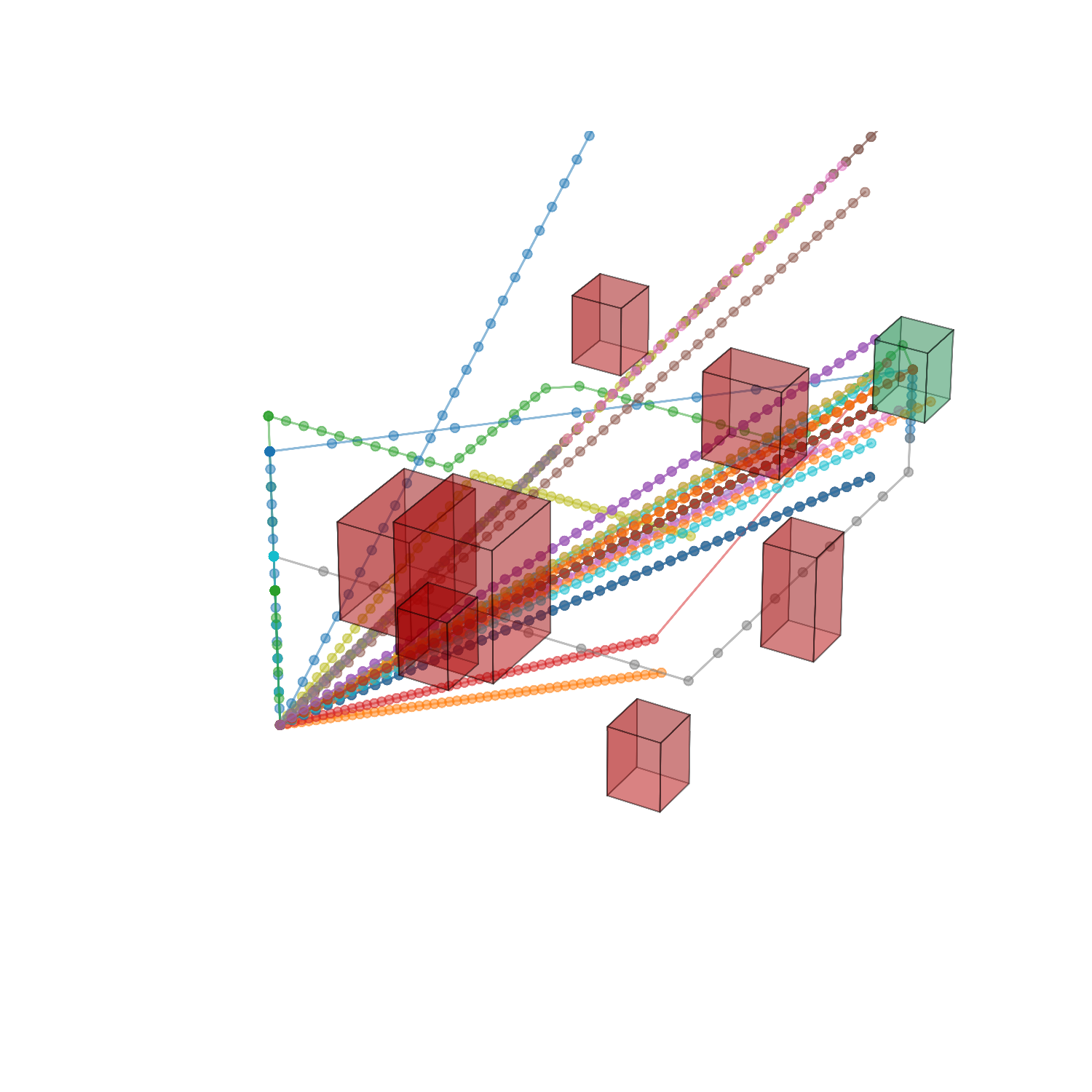}
    \caption{Conventional planner}
    \label{fig:randa_conv}
\end{subfigure}
\caption{Generated paths for R\&A.}
\label{fig:randa_path}
\end{figure}

The robot paths generated from multi-trial experiments are illustrated in Sec.~\ref{fig:randa_path}. It can be seen from Fig.~\ref{fig:randa_conv} that the conventional NL-based planner generates highly variant paths, reflecting a high level of ambiguity and uncertainty of the LLM agent. Moreover, most paths present piece-wise linear shapes, showing the lack of smoothness. Moreover, many paths fail to reach the goal or even collide with the obstacles, leading to failure of the R\&A task. This can also be quantitatively viewed from Tab.~\ref{Tab:randa_Results} that only 51\% of its trajectories ultimately reach the goal and only 36\% successfully avoid collisions. 

On the contrary, the VernaCopter planner successfully achieves goal-reaching and collision avoidance for all trials, as shown in Tab.~\ref{Tab:randa_Results}. Fig.~\ref{fig:randa_vcp} also shows that it generates highly consistent paths. This reflects that the involvement of STL successfully reduces the uncertainty brought up by the LLM agent. Moreover, the generated paths have decent smoothness due to the customized optimizer. This reflects another advantage of the proposed VernaCopter planner that the specific shape of the path is determined by a precise mathematical solver instead of the LLM itself. Since the Vernacopter planner with one-shot prompts already shows decent performance, extending it to a conversation-prompting mode is not necessary for this scenario.


\linespread{1.2}
\begin{table}[htbp]
\centering
\caption{\footnotesize THE PERFORMANCE FOR R\&A.} 
\begin{tabular}{c|c|c}
\hline
Performance & VernaCopter planner & Conventional planner \\
\hline
Goal-Reaching & $100\%$ (50/50) & $51\%$ (23/45) \\
Collision-Free & $100\%$ (50/50) & $36\%$ (16/45) \\
\hline
\end{tabular} 
\label{Tab:randa_Results}
\end{table}
\linespread{1}

\subsection{Case II: Treasure Hunt}

The treasure hunt scenario has been mentioned briefly in Sec.~\ref{sec:intro}. As shown in Fig. \ref{fig:treasure_hunt}, the VernaCopter must reach the treasure \textit{chest} (yellow) in an interdependent order: \textit{key} (green) $\rightarrow$ \textit{door} (blue) $\rightarrow$ \textit{chest}. During the process, VernaCopter cannot leave the room or collide with the walls. This case is an original scenario that typically represents a compact motion planning problem with interdependent subtasks. The robot not only needs to reach several goals but also must follow a certain sequential order. This case can validate whether a semantic-enabled planner can correctly reconstruct the solution space from the user prompts and reason about the best solution.

For both the VernaCopter and the conventional planners, we use the one-shot prompting template presented in Appx.-C, with the task description and generation rule predefined for the PA as the system instruction. The one-shot command is given as ``\textit{Go to the key in the first 30 seconds, then go to the chest. Avoid all walls and stay in the room at all times. The door will open when you reach the key.}". The conventional prompt-based planner is designed the same as the one in Case I. Its PA directly generates the robot path instead of STL specifications. For both planners, we perform 50 trials to inspect whether they generate successful paths.

\begin{figure}[htbp]
\centering
\begin{subfigure}{0.48\linewidth}
    \includegraphics[width=\linewidth, trim={2cm 2cm 2cm 8cm},clip]{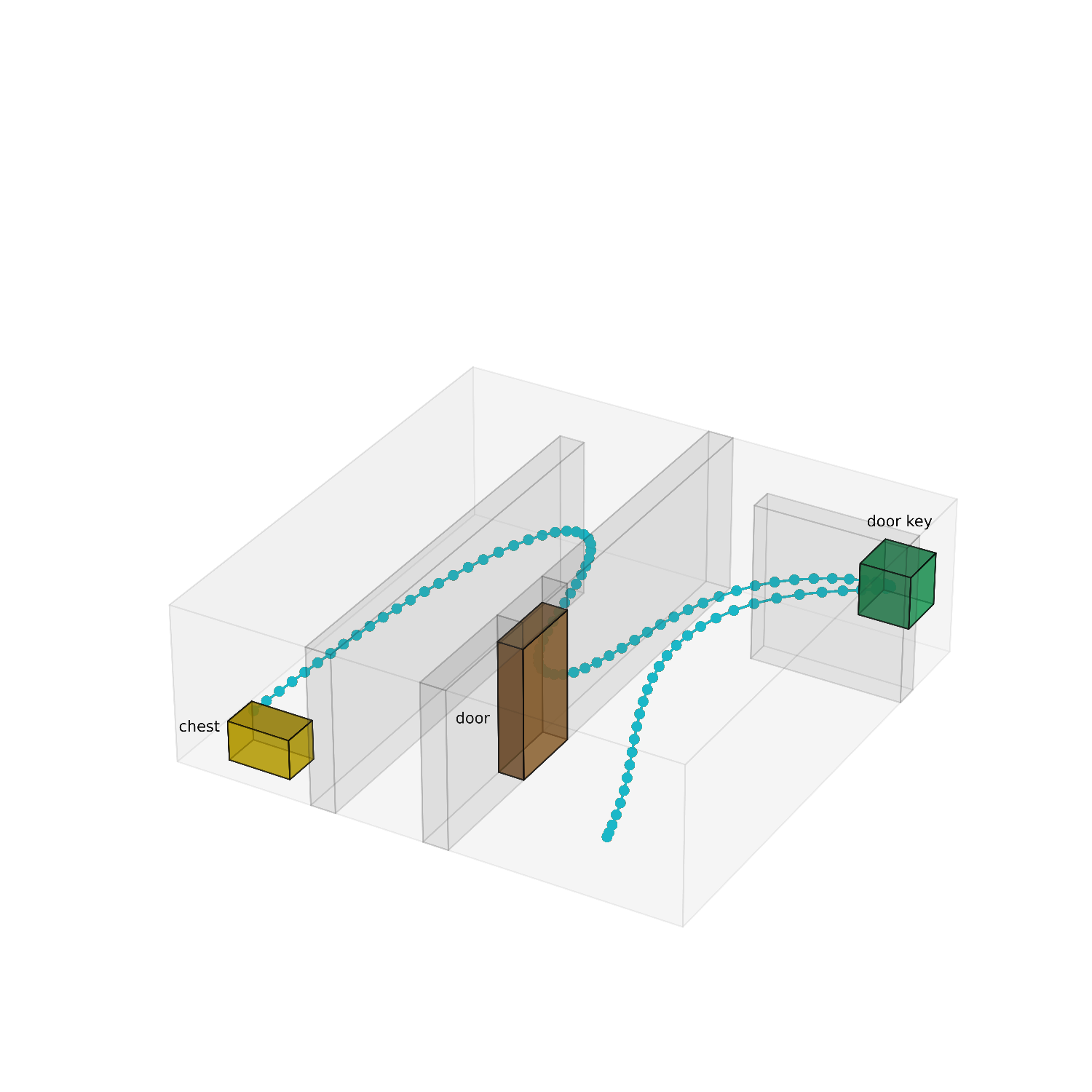}
    \caption{VernaCopter planner}
    \label{fig:th_vcp}
\end{subfigure}
\begin{subfigure}{0.48\linewidth}
    \includegraphics[width=\linewidth, trim={2cm 2cm 2cm 8cm},clip]{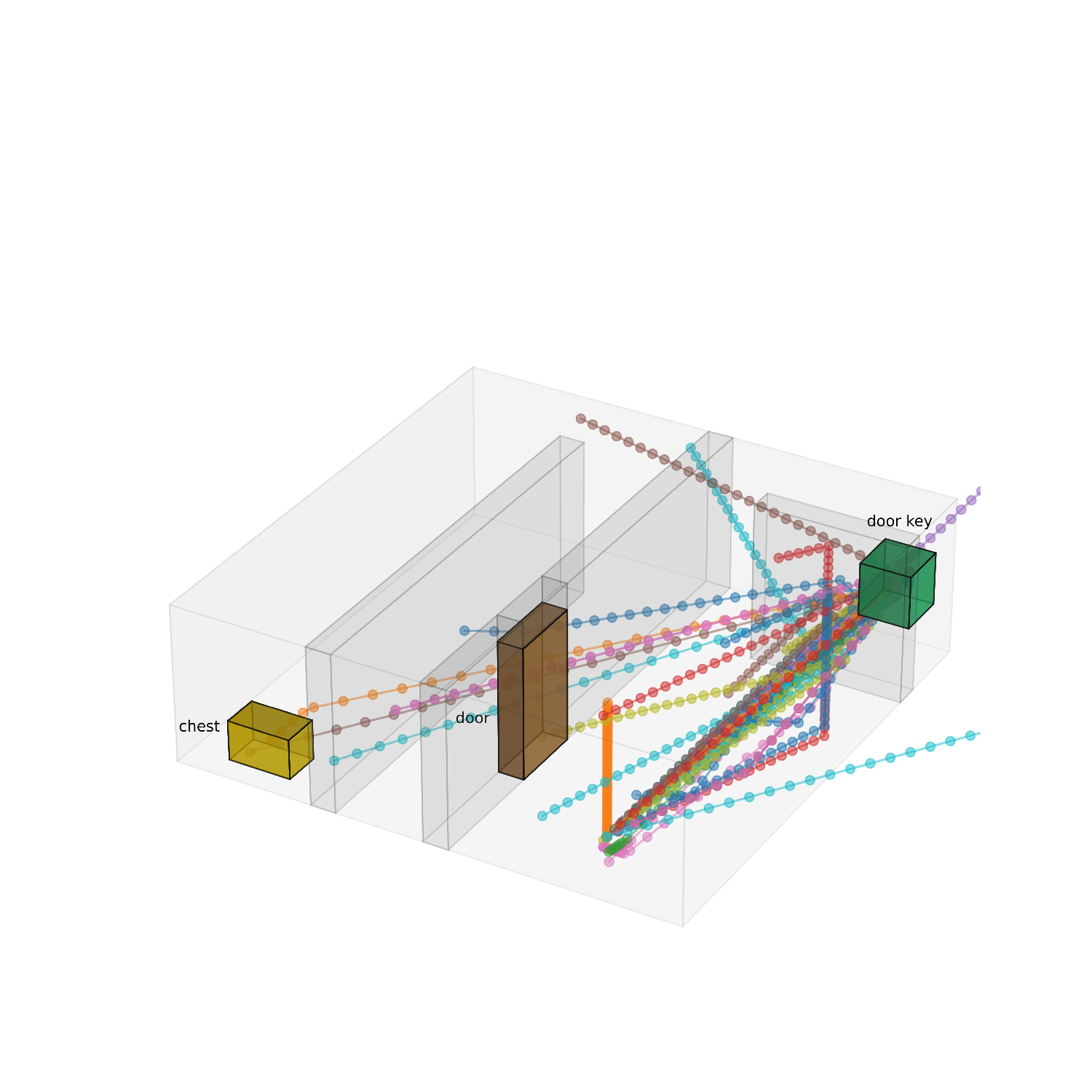}
    \caption{Conventional planner}
    \label{fig:th_conv}
\end{subfigure}
\caption{Generated paths for treasure hunt.}
\label{fig:th_paths}
\end{figure}

The robot paths generated from multi-trial experiments are illustrated in Fig.~\ref{fig:th_paths}, with the quantitative success rate displayed in Tab.~\ref{Tab:th_Results}. It can be seen from Fig.~\ref{fig:th_conv} that the conventional prompt-based planner generates highly variant paths, similar to the results in Sec.~\ref{sec:case1}, reflecting a high level of ambiguity and uncertainty of the LLM agent. Similar non-smoothness is also witnessed in the results since LLM cannot process high-resolution path tuning. Moreover, the distinguishment of the success rates between the planners is more significant. Only 4\% of the trials successfully reach the goal and only 52\% trials are collision-free. Note that a higher collision-free rate than Case I does not imply its higher performance. Instead, it is because Case II has more sparse obstacles. The drastic drop in the goal-reaching rate indicates that a pure language-prompting LLM has difficulties in properly reconstructing the correct logical relations among different subtasks, leading to high a failure rate.

Similar to Case I, the VernaCopter planner successfully achieves goal-reaching and collision avoidance for all trials, with decent smoothness, as shown in Tab.~\ref{Tab:th_Results} and Fig.~\ref{fig:randa_vcp}. This confirms the efficacy of the proposed VernaCopter planner in efficiency and disambiguation, even for complex sequential tasks. In this case, we can see that one-shot prompting already presents a sufficiently decent performance for the VernaCopter planner in the treasure hunt scenario.

\linespread{1.2}
\begin{table}[htbp]
\centering
\caption{\footnotesize THE PERFORMANCE FOR TREASURE HUNT.} 
\begin{tabular}{c|c|c}
\hline
Performance & VernaCopter planner & Conventional planner \\
\hline
Goal-Reaching & $100\%$ (50/50) & $4\%$ (2/50) \\
Collision-Free & $100\%$ (50/50) & $52\%$ (26/50)\\ 
\hline
\end{tabular} 
\label{Tab:th_Results}
\end{table}
\linespread{1}

\section{Discussion}\label{sec:Discussion}

The experimental results in Sec.~\ref{sec:Experimental_studies} reflect an obvious advantage of the proposed formal-specification-enabled VernaCopter planner over a conventional NL-based planner with better generation consistency and higher flexibility in path shape control. The main reason for such an advantage is that the involvement of formal specifications as a bridge between NL and robot control command can effectively reduce the ambiguity in NL and the uncertainty brought up by LLM. Based on this, proper design of prompts is also important to guide the LLM-based PA to generate the desired results. Nevertheless, compared to the conventional prompt-based robot planner, prompting for VernaCopter is easier since it has a more focused scope. Specifically, it only needs to focus on inducing the precise generation of desired formal specifications and let a specific optimizer take care of the quality of the generation results. Here, we have utilized the twofold property of formal specifications. Its precise nature allows for generating consistent results without bringing up additional uncertainty. Meanwhile, its abstract nature has a closer distance to NL than the objective of a specific robot task. From this perspective, we believe formal-language-powered LLM systems are promising to be a practical solution to inspire applicable language-enabled robots.

Although the VernaCopter planner designed in this paper serves as a decent prototype, there are still some limitations that may induce interesting studies in the future. Firstly, the planner may generate infeasible or conflicting specifications, leading to generation failure, even though a feasible path does exist for the given task. The possible reason is due to an ill-defined specification,
where the constraints of the task are too restrictive. On the other hand, this also reflects that the proposed VernaCopter planner does not ensure completeness of the task execution, although sufficiency can be guaranteed thanks to SynCheQ. One possible solution to resolve this issue is to separate or decompose a long and complex specification into shorter or simpler ones~\cite{zhang2023modularized}. In this sense, prior knowledge and heuristics are leveraged to simplify the problem for LLM. Another solution is to fine-tune LLM such that it is more suitable for generating formal specifications, in a sense training a new formal language expert. Previous research \cite{10412086} has investigated fine-tuning pre-trained LLMs to increase output specificity toward the method and syntax used. However, it is difficult to obtain large enough datasets to perform such fine-tuning, and current methods have to do with weak supervision using few-shot learning. Nevertheless, we believe fine-tuning would be worth investigating in the future. LLM-based formal language expert is expected to greatly promote the performance of language-enabled robots.




\vspace{0.3cm}
\section{Conclusion}\label{sec:Conclusions}

This paper proposes a novel LLM-centered robot motion planner leveraged by formal specifications. Compared to the conventional NL-based planner, it ensures high consistency and reliability due to the reduced ambiguity and uncertainty. The current prototype still suffers from conservativeness and possible infeasibility which could be addressed by specification decomposition approaches or LLM fine-tuning. Future work will also look into improving the proposed planner in complicated scenarios with conversational prompts.




\section*{Appendix}

\subsection{Signal Temporal Logic (STL)}\label{sec:stl}

The specification for a real-valued signal can be described using STL. For a discrete-time signal $\mathbf{x}:=x_0x_1\cdots$, where $x_k\in \mathbb{R}^n$ for $k\in \mathbb{N}$ and $n \in \mathbb{Z}^+$, the syntax of STL is recursively defined as~\cite{Maler2004}
\begin{equation}
\varphi::= \, \top \,|\, \mu \,|\, \neg \varphi \,|\, \varphi_1 \wedge \varphi_2 \,|\, \notag \varphi_1 \mathsf{U}_{[k_1, k_2]} \varphi_2,
\end{equation}
where $\varphi$, $\varphi_1$, and $\varphi_2$ are STL formulas, $\mu$ is a predicate $\mu:=\left\{\begin{array}{ll}\top, & h(x_k) \geq 0 \\ \bot, & h(x_k)<0\end{array}\right.$ evaluating a function $h: \mathbb{R}^n \rightarrow \mathbb{R}$, $k \in \mathbb{N}$, $\lnot$ and $\wedge$ are the \textit{negation} and the \textit{conjunction} operators respectively, and $\mathsf{U}_{[\,a,b\,]}$ is the \textit{until} operator associated with a time interval $[\,k_1,k_2\,]$, with $k_1,k_2 \in \mathbb{N}$ and $k_1 \leq k_2$. The \textit{disjunction}, \textit{eventually}, and \textit{always} operators are defined as $\varphi_1 \!\vee\! \varphi_2 \!:=\! \lnot \left(\lnot \varphi_1 \wedge \lnot \varphi_2 \right)$, $\mathsf{F}_{[\,a,\,b\,]} \varphi \!:=\! \top \mathsf{U}_{[\,a,\,b\,]} \varphi$, and $\mathsf{G}_{[\,a,\,b\,]} \varphi\!:=\! \lnot \left( \top \mathsf{U}_{[\,a,\,b\,]} \lnot \varphi \right)$. 

The satisfaction of an STL formula $\varphi$ given a signal $\textbf{x}$ and time $k$ is denoted by $(\textbf{x}, k) \vDash \varphi$ and given by the following recursively defined semantics.
\begin{align*}
(\textbf{x}, k) &\vDash \neg \varphi \Leftrightarrow \neg ((\textbf{x}, k) \vDash \varphi), \\
(\textbf{x}, k) &\vDash \varphi_1 \wedge \varphi_2 \Leftrightarrow (\textbf{x}, k) \vDash \varphi_1 \wedge (\textbf{x}, k) \vDash \varphi_2, \\
(\textbf{x}, k) &\vDash \varphi_1 \vee \varphi_2 \Leftrightarrow (\textbf{x}, k) \vDash \varphi_1 \vee (\textbf{x}, k) \vDash \varphi_2, \\
(\textbf{x}, k) &\vDash \mathsf{F}_{[k_1, k_2]} \varphi \Leftrightarrow  \exists\, k' \in [k+k_1, k+k_2], \text{ s.t. } (\textbf{x}, k') \vDash \varphi, \\
(\textbf{x}, k) &\vDash \mathsf{G}_{[k_1, k_2]} \varphi \Leftrightarrow  \forall\, k' \in [k+k_1, k+k_2], \text{ s.t. } (\textbf{x}, k') \vDash \varphi, \\
(\textbf{x}, k) &\vDash \varphi_1 \mathsf{U}_{[k_1, k_2]} \varphi_2 \Leftrightarrow \exists\, k' \in [k+k_1, k+k_2],  \\
& \quad \quad \quad \text{ s.t. } (\textbf{x}, k') \vDash \varphi \text{ and } \forall\, k'' \in [k, k'], (\textbf{x}, k'') \vDash \varphi.
\end{align*}

For $k=0$, the symbol $k$ can be omitted, rendering  $\mathbf{x} \vDash \varphi$.
The robustness of an STL formula $\varphi$ denoted as $\rho^{\varphi}(\mathbf{x},k)$, with $\rho^{\varphi}(\mathbf{x},k)>0 \leftrightarrow (\mathbf{x}, k) \vDash \varphi$, is inductively defined as
\begin{equation*}
\begin{split}
&\textstyle \rho^{\mu}(\mathbf{x},k) \!:=\! h(x_k),~
\rho^{\lnot \varphi}(\mathbf{x},k) \!:=\! - \rho^{\varphi}(\mathbf{x},k),\\
&\textstyle \rho^{\varphi_1 \wedge \varphi_2}(\mathbf{x},k)\!: =\! \min(\rho^{\varphi_1}(\mathbf{x}, k),\, \rho^{\varphi_2}(\mathbf{x},k)), \\
&\textstyle \rho^{\varphi_1 \vee \varphi_2}(\mathbf{x},k) \!: =\! \max(\rho^{\varphi_1}(\mathbf{x},k),\, \rho^{\varphi_2}(\mathbf{x},k)), \\
&\textstyle \rho^{\mathsf{F}_{[\,a,\,b\,]} \varphi}(\mathbf{x}, k ) \!:=\! \max_{k' \in [\,k+a,\,k+b\,]} \rho^{\varphi}(\mathbf{x}, k'), \\
&\textstyle \rho^{\mathsf{G}_{[\,a,\,b\,]} \varphi}(\mathbf{x},k)\!:=\min_{k' \in [\,k+a,\,k+b\,]} \rho^{\varphi}(\mathbf{x}, k'),\\
&\textstyle \rho^{\varphi_1 \!\mathsf{U}_{[\,a,\,b\,]}\! \varphi_2\!}(\mathbf{x},k) := \max_{k' \in [\,k+a,\,k+b\,]} 
(\min ( \rho^{\varphi_2}(\mathbf{x}, k'),\\
&\textstyle ~~~~~~~~~~~~~~~~~~~~~~~~~~\min_{k'' \in [\,k, \,k'\,]} \rho^{\varphi_1}(\mathbf{x}, k'') ) ).
\end{split}
\end{equation*}
For $k=0$, the robustness reads $\rho^{\varphi}(\mathbf{x})$.

\subsection{System Models and Motion Planning}\label{sec:Drone_model}

\begin{figure}[!t]
\vspace{0.2cm}
    \centering
    \includegraphics[width=0.98\linewidth]{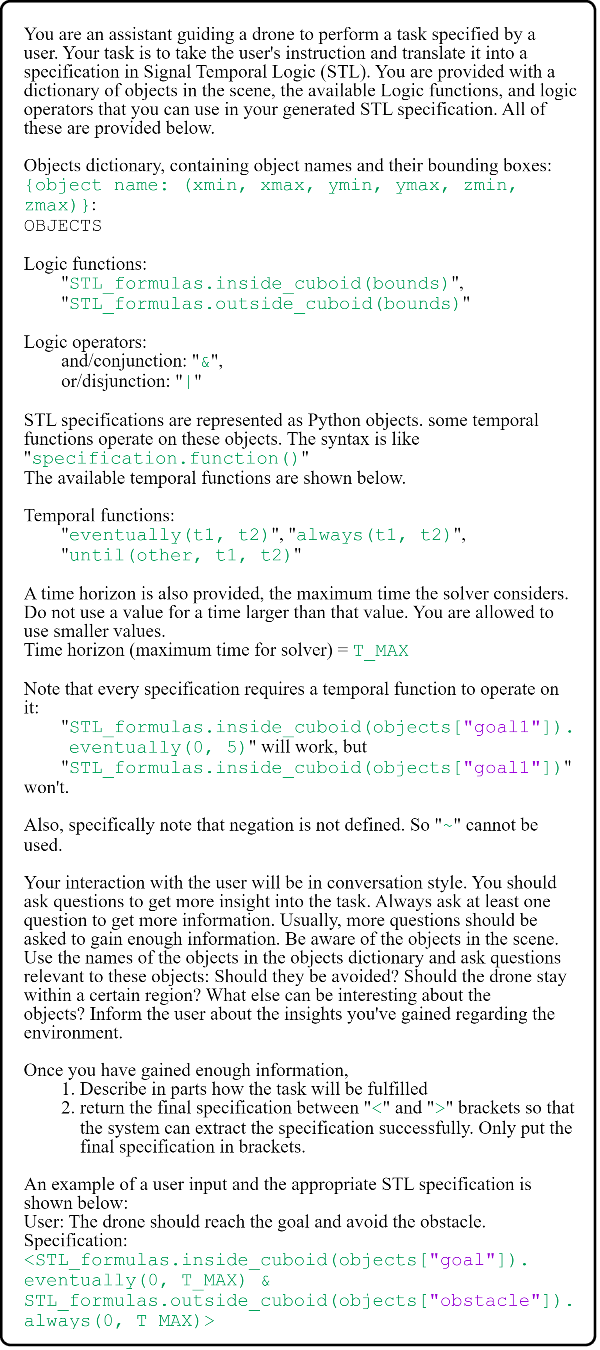}
    \caption{The one-shot prompting instruction.}
    \vspace{-0.4cm}
    \label{fig:GPT_planner_instructions}
\end{figure}

The VernaCopter is described as a double-integrator model,
\begin{equation}\label{eq:model}
\left\{
\begin{array}{l}
    p_{t+1} = p_t + \Delta t v_t, \\
    v_{t+1} = v_t + \Delta t a_t
\end{array}
\right.
\end{equation}
where $p_t, v_t, a_t \in \mathbb{R}^3$ are the position, linear velocity, and acceleration of the VernaCopter at time $t\in \mathbb{N}$, and $\Delta t \in \mathbb{R}^+$ is the discrete sampling time. Let us denote its path as a signal $\mathbf{x}:=(p_0,v_0)(p_1,v_1)\cdots(p_T,v_T)$ within a finite control horizon $T \in \mathbb{Z}^+$. A motion planning task can be specified using an STL formula $\varphi$ as defined in Sec.~\ref{sec:stl}, with a robustness constraint $\rho^{\varphi}(\mathbf{x}) > 0$. Moreover, the motion planning task usually requires minimizing certain cost, typically $J(\mathbf{x}, \mathbf{u}):= \sum_{t=0}^T \!\left(v_t^{\top} Q v_t + a_t^{\top} R a_t \right)$, where $\mathbf{u}:=a_0a_1\cdots a_T$ is the control sequence and $Q, R \in \mathbb{R}^{3 \times 3}$ are cost matrices. Then, motion planning can be formulated as the following optimization problem with a given initial condition $p_0,v_0\in \mathbb{R}^3$,
\begin{equation*}
\begin{split}
 &\min_{\mathbf{u}} J(\mathbf{x}, \mathbf{u}) \\
\text{s.t. } & \eqref{eq:model}, \forall\, t=0,1,\cdots,T-1, \\
& \rho^{\varphi} (\mathbf{x}) \geq 0, \\
& p_{\min} \leq p_t \leq p_{\max},~v_{\min} \leq v_t \leq v_{\max},\\
& a_{\min} \leq a_t \leq a_{\max},~\forall\, t=0,1,\cdots,T, 
\end{split}
\end{equation*}
where $p_{\{\min,\max\}}$, $v_{\{\min,\max\}}$, and $a_{\{\min,\max\}}$ are the limits of position, velocity, and control input of VernaCopter. This problem can be converted to a Mixed-Integer Convex Programming (MICP) problem and solved using off-the-shelf tools \cite{kurtz2022mixed}.

\subsection{Prompting Templates for the VernaCopter Planner}\label{sec:template}

The prompting instructions for the VernaCopter planner are illustrated in Fig.~\ref{fig:GPT_planner_instructions}. The objects and regions in the planning space are represented with rectangular cuboidal bounding boxes. The drone can be inside or outside the bounding box of a specified region, denoted with STL formulas denoted as
\textit{\small STL\_formulas.inside\_cuboid(bounds)} and \textit{\small STL\_formulas.outside\_}
\textit{\small cuboid(bounds)}, respectively. Temporal operators operate on STL formulas (denoted with $\pi$): \textit{\small $\pi$.eventually(t1, t2)} is true whenever $\pi$ holds \textit{at any time} between times $t1$ and $t2$; \textit{\small $\pi$.always(t1, t2)} is true whenever $\pi$ holds \textit{at all times} between times $t1$ and $t2$; and \textit{\small $\pi$.until(other, t1, t2)} is true whenever $\pi$ holds at all times before \textit{other} holds between times $t1$ and $t2$. These definitions are in line with Section \ref{sec:stl}.

\bibliographystyle{IEEEtran}
\bibliography{IEEEabrv, ref.bib}

\begin{thebibliography}{10}
\providecommand{\url}[1]{#1}
\csname url@samestyle\endcsname
\providecommand{\newblock}{\relax}
\providecommand{\bibinfo}[2]{#2}
\providecommand{\BIBentrySTDinterwordspacing}{\spaceskip=0pt\relax}
\providecommand{\BIBentryALTinterwordstretchfactor}{4}
\providecommand{\BIBentryALTinterwordspacing}{\spaceskip=\fontdimen2\font plus
\BIBentryALTinterwordstretchfactor\fontdimen3\font minus \fontdimen4\font\relax}
\providecommand{\BIBforeignlanguage}[2]{{%
\expandafter\ifx\csname l@#1\endcsname\relax
\typeout{** WARNING: IEEEtran.bst: No hyphenation pattern has been}%
\typeout{** loaded for the language `#1'. Using the pattern for}%
\typeout{** the default language instead.}%
\else
\language=\csname l@#1\endcsname
\fi
#2}}
\providecommand{\BIBdecl}{\relax}
\BIBdecl

\bibitem{ahn2018interactive}
H.~Ahn, S.~Choi, N.~Kim, G.~Cha, and S.~Oh, ``Interactive text2pickup networks for natural language-based human--robot collaboration,'' \emph{IEEE Robotics and Automation Letters}, vol.~3, no.~4, pp. 3308--3315, 2018.

\bibitem{zemni2024comparative}
B.~Zemni, M.~Zitouni, F.~Bouhadiba, and M.~Almutairi, ``A comparative study of the lexical ambiguity of arabic, english, and french in natural language processing,'' \emph{Journal of Intercultural Communication}, vol.~24, no.~1, pp. 203--203, 2024.

\bibitem{kahuttanaseth2018commanding}
W.~Kahuttanaseth, A.~Dressler, and C.~Netramai, ``Commanding mobile robot movement based on natural language processing with rnn encoderdecoder,'' in \emph{2018 5th International Conference on Business and Industrial Research (ICBIR)}.\hskip 1em plus 0.5em minus 0.4em\relax IEEE, 2018, pp. 161--166.

\bibitem{johri2021natural}
P.~Johri, S.~K. Khatri, A.~T. Al-Taani, M.~Sabharwal, S.~Suvanov, and A.~Kumar, ``Natural language processing: History, evolution, application, and future work,'' in \emph{Proceedings of 3rd International Conference on Computing Informatics and Networks: ICCIN 2020}.\hskip 1em plus 0.5em minus 0.4em\relax Springer, 2021, pp. 365--375.

\bibitem{gpt3}
T.~B. Brown, B.~Mann, N.~Ryder \emph{et~al.}, ``Language models are few-shot learners,'' 2020.

\bibitem{geminiteam2023gemini}
G.~Team, R.~Anil, S.~Borgeaud \emph{et~al.}, ``Gemini: A family of highly capable multimodal models,'' 2023.

\bibitem{naveed2024comprehensive}
H.~Naveed, A.~U. Khan, S.~Qiu, M.~Saqib, S.~Anwar, M.~Usman, N.~Akhtar, N.~Barnes, and A.~Mian, ``A comprehensive overview of large language models,'' 2024.

\bibitem{xu2024survey}
M.~Xu, W.~Yin, D.~Cai, R.~Yi, D.~Xu, Q.~Wang, B.~Wu, Y.~Zhao, C.~Yang, S.~Wang \emph{et~al.}, ``A survey of resource-efficient llm and multimodal foundation models,'' \emph{arXiv preprint arXiv:2401.08092}, 2024.

\bibitem{zhang2024llm}
Y.~Zhang, S.~Mao, T.~Ge, X.~Wang, A.~de~Wynter, Y.~Xia, W.~Wu, T.~Song, M.~Lan, and F.~Wei, ``Llm as a mastermind: A survey of strategic reasoning with large language models,'' \emph{arXiv preprint arXiv:2404.01230}, 2024.

\bibitem{vemprala2023chatgpt}
S.~Vemprala, R.~Bonatti, A.~Bucker, and A.~Kapoor, ``Chatgpt for robotics: Design principles and model abilities,'' 2023.

\bibitem{Chen2023}
Y.~Chen, J.~Arkin, C.~Dawson, Y.~Zhang, N.~Roy, and C.~Fan, ``Autotamp: Autoregressive task and motion planning with llms as translators and checkers,'' 2024.

\bibitem{wang2024llm}
R.~Wang, Z.~Yang, Z.~Zhao, X.~Tong, Z.~Hong, and K.~Qian, ``Llm-based robot task planning with exceptional handling for general purpose service robots,'' \emph{arXiv preprint arXiv:2405.15646}, 2024.

\bibitem{arawjo2024chainforge}
I.~Arawjo, C.~Swoopes, P.~Vaithilingam, M.~Wattenberg, and E.~L. Glassman, ``Chainforge: A visual toolkit for prompt engineering and llm hypothesis testing,'' in \emph{Proceedings of the CHI Conference on Human Factors in Computing Systems}, 2024, pp. 1--18.

\bibitem{zhao2024survey}
Z.~Zhao, S.~Cheng, Y.~Ding, Z.~Zhou, S.~Zhang, D.~Xu, and Y.~Zhao, ``A survey of optimization-based task and motion planning: From classical to learning approaches,'' 2024.

\bibitem{liu2024uncertainty}
L.~Liu, Y.~Pan, X.~Li, and G.~Chen, ``Uncertainty estimation and quantification for llms: A simple supervised approach,'' \emph{arXiv preprint arXiv:2404.15993}, 2024.

\bibitem{gao2023ambiguity}
L.~Gao, A.~Chaudhary, K.~Srinivasan, K.~Hashimoto, K.~Raman, and M.~Bendersky, ``Ambiguity-aware in-context learning with large language models,'' \emph{arXiv preprint arXiv:2309.07900}, 2023.

\bibitem{wake2023chatgpt}
N.~Wake, A.~Kanehira, K.~Sasabuchi, J.~Takamatsu, and K.~Ikeuchi, ``Chatgpt empowered long-step robot control in various environments: A case application,'' \emph{arXiv preprint arXiv:2304.03893}, 2023.

\bibitem{valmeekam2022large}
K.~Valmeekam, A.~Olmo, S.~Sreedharan, and S.~Kambhampati, ``Large language models still can't plan (a benchmark for llms on planning and reasoning about change),'' \emph{arXiv preprint arXiv:2206.10498}, 2022.

\bibitem{liu2024enhancing}
H.~Liu, Y.~Zhu, K.~Kato, A.~Tsukahara, I.~Kondo, T.~Aoyama, and Y.~Hasegawa, ``Enhancing the llm-based robot manipulation through human-robot collaboration,'' \emph{arXiv preprint arXiv:2406.14097}, 2024.

\bibitem{singh2023progprompt}
I.~Singh, V.~Blukis, A.~Mousavian, A.~Goyal, D.~Xu, J.~Tremblay, D.~Fox, J.~Thomason, and A.~Garg, ``Progprompt: Generating situated robot task plans using large language models,'' in \emph{International Conference on Robotics and Automation (ICRA)}, 2023, available: \url{https://arxiv.org/abs/2209.11302}.

\bibitem{ding2023task}
Y.~Ding, X.~Zhang, C.~Paxton, and S.~Zhang, ``Task and motion planning with large language models for object rearrangement,'' \emph{arXiv preprint arXiv:2303.06247}, 2023.

\bibitem{10412086}
Y.~Jin, D.~Li, Y.~A, J.~Shi, P.~Hao, F.~Sun, J.~Zhang, and B.~Fang, ``Robotgpt: Robot manipulation learning from chatgpt,'' \emph{IEEE Robotics and Automation Letters}, vol.~9, no.~3, pp. 2543--2550, 2024.

\bibitem{rana2023sayplan}
\BIBentryALTinterwordspacing
K.~Rana, J.~Haviland, S.~Garg, J.~Abou-Chakra, I.~Reid, and N.~Suenderhauf, ``Sayplan: Grounding large language models using 3d scene graphs for scalable task planning,'' in \emph{7th Annual Conference on Robot Learning}, 2023. [Online]. Available: \url{https://openreview.net/forum?id=wMpOMO0Ss7a}
\BIBentrySTDinterwordspacing

\bibitem{huang2022language}
W.~Huang, P.~Abbeel, D.~Pathak, and I.~Mordatch, ``Language models as zero-shot planners: Extracting actionable knowledge for embodied agents,'' 2022.

\bibitem{ahn2022do}
M.~Ahn, A.~Brohan, N.~Brown, Y.~Chebotar, O.~Cortes, B.~David, C.~Finn, K.~Gopalakrishnan, K.~Hausman, A.~Herzog \emph{et~al.}, ``Do as i can, not as i say: Grounding language in robotic affordances,'' \emph{arXiv preprint arXiv:2204.01691}, 2022.

\bibitem{lin2023text2motion}
K.~Lin, C.~Agia, T.~Migimatsu, M.~Pavone, and J.~Bohg, ``Text2motion: From natural language instructions to feasible plans,'' \emph{arXiv preprint arXiv:2303.12153}, 2023.

\bibitem{wu2023tidybot}
J.~Wu, R.~Antonova, A.~Kan, M.~Lepert, A.~Zeng, S.~Song, J.~Bohg, S.~Rusinkiewicz, and T.~Funkhouser, ``Tidybot: Personalized robot assistance with large language models,'' \emph{arXiv preprint arXiv:2305.05658}, 2023.

\bibitem{huang2022inner}
W.~Huang, F.~Xia, T.~Xiao, H.~Chan, J.~Liang, P.~Florence, A.~Zeng, J.~Tompson, I.~Mordatch, Y.~Chebotar \emph{et~al.}, ``Inner monologue: Embodied reasoning through planning with language models,'' \emph{arXiv preprint arXiv:2207.05608}, 2022.

\bibitem{Bucker2022}
A.~Bucker, L.~Figueredo, S.~Haddadin, A.~Kapoor, S.~Ma, S.~Vemprala, and R.~Bonatti, ``Latte: Language trajectory transformer,'' in \emph{2023 IEEE International Conference on Robotics and Automation (ICRA)}.\hskip 1em plus 0.5em minus 0.4em\relax IEEE, 2023, pp. 7287--7294.

\bibitem{kloetzer2007temporal}
M.~Kloetzer and C.~Belta, ``Temporal logic planning and control of robotic swarms by hierarchical abstractions,'' \emph{IEEE Transactions on Robotics}, vol.~23, no.~2, pp. 320--330, 2007.

\bibitem{lindemann2019coupled}
L.~Lindemann, J.~Nowak, L.~Sch{\"o}nb{\"a}chler, M.~Guo, J.~Tumova, and D.~V. Dimarogonas, ``Coupled multi-robot systems under linear temporal logic and signal temporal logic tasks,'' \emph{IEEE Transactions on Control Systems Technology}, vol.~29, no.~2, pp. 858--865, 2019.

\bibitem{qi23cdc}
S.~Qi, Z.~Zhang, S.~Haesaert, and Z.~Sun, ``Automated formation control synthesis from temporal logic specifications,'' in \emph{2023 62nd IEEE Conference on Decision and Control (CDC)}, 2023, pp. 5165--5170.

\bibitem{white2023prompt}
J.~White, Q.~Fu, S.~Hays, M.~Sandborn, C.~Olea, H.~Gilbert, A.~Elnashar, J.~Spencer-Smith, and D.~C. Schmidt, ``A prompt pattern catalog to enhance prompt engineering with chatgpt,'' 2023.

\bibitem{ekin2023prompt}
S.~Ekin, ``Prompt engineering for chatgpt: A quick guide to techniques, tips, and best practices,'' \emph{TechRxiv}, 2023, e-Prints posted on TechRxiv are preliminary reports that are not peer-reviewed. They should not be regarded as conclusive, guide clinical practice/health-related behavior, or be reported in the media as established information.

\bibitem{shi2024yell}
L.~X. Shi, Z.~Hu, T.~Z. Zhao, A.~Sharma, K.~Pertsch, J.~Luo, S.~Levine, and C.~Finn, ``Yell at your robot: Improving on-the-fly from language corrections,'' \emph{arXiv preprint arXiv:2403.12910}, 2024.

\bibitem{liu2024leveraging}
X.~Liu, P.~Li, W.~Yang, D.~Guo, and H.~Liu, ``Leveraging large language model for heterogeneous ad hoc teamwork collaboration,'' \emph{arXiv preprint arXiv:2406.12224}, 2024.

\bibitem{yuan2024rag}
J.~Yuan, S.~Sun, D.~Omeiza, B.~Zhao, P.~Newman, L.~Kunze, and M.~Gadd, ``Rag-driver: Generalisable driving explanations with retrieval-augmented in-context learning in multi-modal large language model,'' \emph{arXiv preprint arXiv:2402.10828}, 2024.

\bibitem{gu2023systematic}
J.~Gu, Z.~Han, S.~Chen, A.~Beirami, B.~He, G.~Zhang, R.~Liao, Y.~Qin, V.~Tresp, and P.~Torr, ``A systematic survey of prompt engineering on vision-language foundation models,'' 2023.

\bibitem{wei2023chainofthought}
J.~Wei, X.~Wang, D.~Schuurmans, M.~Bosma, B.~Ichter, F.~Xia, E.~Chi, Q.~Le, and D.~Zhou, ``Chain-of-thought prompting elicits reasoning in large language models,'' 2023.

\bibitem{sun2024determlr}
H.~Sun, W.~Xu, W.~Liu, J.~Luan, B.~Wang, S.~Shang, J.-R. Wen, and R.~Yan, ``Determlr: Augmenting llm-based logical reasoning from indeterminacy to determinacy,'' in \emph{Proceedings of the 62nd Annual Meeting of the Association for Computational Linguistics (Volume 1: Long Papers)}, 2024, pp. 9828--9862.

\bibitem{Plaku2015MotionPW}
\BIBentryALTinterwordspacing
E.~Plaku and S.~Karaman, ``Motion planning with temporal-logic specifications: Progress and challenges,'' \emph{AI Commun.}, vol.~29, pp. 151--162, 2015. [Online]. Available: \url{https://api.semanticscholar.org/CorpusID:30347236}
\BIBentrySTDinterwordspacing

\bibitem{da2016formal}
R.~R. da~Silva, B.~Wu, and H.~Lin, ``Formal design of robot integrated task and motion planning,'' in \emph{2016 IEEE 55th Conference on Decision and Control (CDC)}.\hskip 1em plus 0.5em minus 0.4em\relax IEEE, 2016, pp. 6589--6594.

\bibitem{Xie2023}
Y.~Xie, C.~Yu, T.~Zhu, J.~Bai, Z.~Gong, and H.~Soh, ``Translating natural language to planning goals with large-language models,'' \emph{arXiv preprint arXiv:2302.05128}, 2023.

\bibitem{Liu2023}
B.~Liu, Y.~Jiang, X.~Zhang, Q.~Liu, S.~Zhang, J.~Biswas, and P.~Stone, ``Llm+ p: Empowering large language models with optimal planning proficiency,'' \emph{arXiv preprint arXiv:2304.11477}, 2023.

\bibitem{8972130}
I.~Buzhinsky, ``Formalization of natural language requirements into temporal logics: a survey,'' in \emph{2019 IEEE 17th International Conference on Industrial Informatics (INDIN)}, vol.~1, 2019, pp. 400--406.

\bibitem{openai2024gpt4}
J.~Achiam, S.~Adler, S.~Agarwal, L.~Ahmad, I.~Akkaya, F.~L. Aleman, D.~Almeida, J.~Altenschmidt, S.~Altman, S.~Anadkat \emph{et~al.}, ``Gpt-4 technical report,'' 2023.

\bibitem{kurtz2022mixed}
V.~Kurtz and H.~Lin, ``Mixed-integer programming for signal temporal logic with fewer binary variables,'' \emph{IEEE Control Systems Letters}, 2022.

\bibitem{panerati2021learning}
J.~Panerati, H.~Zheng, S.~Zhou, J.~Xu, A.~Prorok, and A.~P. Schoellig, ``Learning to fly---a gym environment with pybullet physics for reinforcement learning of multi-agent quadcopter control,'' in \emph{2021 IEEE/RSJ International Conference on Intelligent Robots and Systems (IROS)}, 2021, pp. 7512--7519.

\bibitem{1606.01540}
G.~Brockman, V.~Cheung, L.~Pettersson, J.~Schneider, J.~Schulman, J.~Tang, and W.~Zaremba, ``Openai gym,'' 2016.

\bibitem{paranjape2015motion}
A.~A. Paranjape, K.~C. Meier, X.~Shi, S.-J. Chung, and S.~Hutchinson, ``Motion primitives and 3d path planning for fast flight through a forest,'' \emph{The International Journal of Robotics Research}, vol.~34, no.~3, pp. 357--377, 2015.

\bibitem{zhang2023modularized}
Z.~Zhang and S.~Haesaert, ``Modularized control synthesis for complex signal temporal logic specifications,'' in \emph{2023 62nd IEEE Conference on Decision and Control (CDC)}.\hskip 1em plus 0.5em minus 0.4em\relax IEEE, 2023, pp. 7856--7861.

\bibitem{Maler2004}
O.~Maler and D.~Nickovic, ``Monitoring temporal properties of continuous signals,'' in \emph{International symposium on formal techniques in real-time and fault-tolerant systems}.\hskip 1em plus 0.5em minus 0.4em\relax Springer, 2004, pp. 152--166.

\end{thebibliography}
\balance

\end{document}